\pdfoutput=1
\documentclass[journal]{IEEEtran}
\IEEEoverridecommandlockouts
 
\usepackage{multirow,tabularx}
\usepackage{multicol}
\usepackage{adjustbox}
\usepackage{booktabs}
\usepackage[labelsep=period]{caption}   
\usepackage{placeins}

\usepackage{hyperref}
\usepackage{tikz}
\usepackage{algorithm}
\usepackage{algorithmic}
\usepackage{cite}
\usepackage{amsmath,amssymb,amsfonts}
\usepackage{graphicx}
\usepackage{textcomp}
\usepackage{xcolor}
\usepackage{rotating}
\usepackage{svg}
\usepackage{subcaption}
\usepackage{gensymb}

\begin{document}

\title{Learning crop type mapping from regional label proportions in large-scale SAR and optical imagery}

\author{Laura~E.C.~La~Rosa,
	Dario~A.B.~Oliveira, and
	Pedram Ghamisi
	\thanks{Laura~E.C.~La~Rosa is with the Dept. of Electrical Engineering, Pontifical Catholic University of Rio de Janeiro, 22451-900, Brazil, and also with the Helmholtz-Zentrum Dresden-Rossendorf (HZDR), Helmholtz Institute Freiberg for Resource Technology, 09599 Freiberg, Germany (e-mails: lauracuerosa@gmail.com, cuela62@hzdr.de).}
	\thanks{Dario~A.B.~Oliveira is with the Data Science in Earth Observation, Technical University of Munich (TUM), Munich, Germany and with the School of Applied Mathematics, Getulio Vargas Foundation, Rio de Janeiro, Brazil. (e-mail: darioaugusto@gmail.com)}
	\thanks{Pedram Ghamisi is with the Helmholtz-Zentrum Dresden-Rossendorf (HZDR), Helmholtz Institute Freiberg for Resource Technology, 09599 Freiberg, Germany, and is also with the Institute of Advanced Research in Artificial Intelligence (IARAI), 1030 Vienna, Austria (e-mail: p.ghamisi@gmail.com).}}

 
\maketitle
\IEEEpeerreviewmaketitle

\begin{abstract}
The application of deep learning algorithms to Earth observation (EO) in recent years has enabled substantial progress in fields that rely on remotely sensed data. However, given the data scale in EO, creating large datasets with pixel-level annotations by experts is expensive and highly time-consuming. In this context, priors are seen as an attractive way to alleviate the burden of manual labeling when training deep learning methods for EO. For some applications, those priors are readily available. Motivated by the great success of contrastive-learning methods for self-supervised feature representation learning in many computer-vision tasks, this study proposes an online deep clustering method using crop label proportions as priors to learn a sample-level classifier based on government crop-proportion data for a whole agricultural region. We evaluate the method using two large datasets from two different agricultural regions in Brazil. Extensive experiments demonstrate that the method is robust to different data types (synthetic-aperture radar and optical images), reporting higher accuracy values considering the major crop types in the target regions. Thus, it can alleviate the burden of large-scale image annotation in EO applications. 

\end{abstract}


\section{Introduction}\label{sec:introduction}

The free availability of the continuously growing body of Earth observation (EO) images has created an urgent need to develop effective and efficient algorithms capable of processing vast amounts of data and extracting crucial information for various applications, ranging from environmental monitoring to precision farming. For many such applications, classification maps need to be derived from the acquired images to identify the underlying processes observed remotely through the images. However, the statistical properties of EO images, such as heterogeneity in space and time, noise in the observed geophysical processes, and the spatial and spectral redundancy, make the automatic classification of EO data a very challenging task \cite{romero2016unsupervised}. 

In this context, deep learning (DL) algorithms have proved to be effective for learning relevant non-linear features directly from data and have become the state-of-art in many EO image analysis tasks, including land use and land cover classification, object detection, change detection, and domain adaptation \cite{ma2019deep,ghanbari2021meta,camps2021deep}. However, training robust DL solutions with good generalization usually requires large amounts of annotated samples. Hence, the current bottleneck in DL for EO is not much about the amount of available data but rather the capability and cost to process and annotate it since pixel-level EO data expert labeling is expensive and time-consuming. Such a challenging scenario creates excellent opportunities for unsupervised and weakly supervised learning solutions. 

Unsupervised DL approaches are mostly based on autoencoders \cite{hinton1994autoencoders} and generative adversarial networks (GANs) \cite{goodfellow2014generative}, with application to a wide range of remote-sensing (RS) tasks, including feature extraction \cite{zhou2015high,zabalza2016novel}, image classification \cite{lv2017remote,othman2016using}, and change detection \cite{gong2017generative}. More recently, contrastive-learning methods have closed the gap between supervised and unsupervised representation learning in many computer-vision applications \cite{chen2020simple,caron2020unsupervised,chen2021exploring}. Those methods promote self-supervised representation learning by enforcing consistency between different augmented views of the same input image via a contrastive loss function applied in the feature space. They have very quickly become popular in the search for meaningful feature representations in EO \cite{ayush2021geography,stojnic2021self,cai2021large}. One of the most popular methods for self-supervised learning using contrast is \textit{Swapping Assignments between Multiple Views} (SwAV) \cite{caron2020unsupervised}, which jointly performs self-supervised feature learning and clustering by combining contrastive learning and optimal transport theory. In SwAV, the optimal transport solver assigns samples to cluster prototypes (or centroids) and computes pseudo-cluster labels to guide the clustering process using an equipartition constraint that ensures that all samples are equally partitioned among the clusters.    

\begin{figure*}[t]
\centering
\includegraphics[scale=0.44]{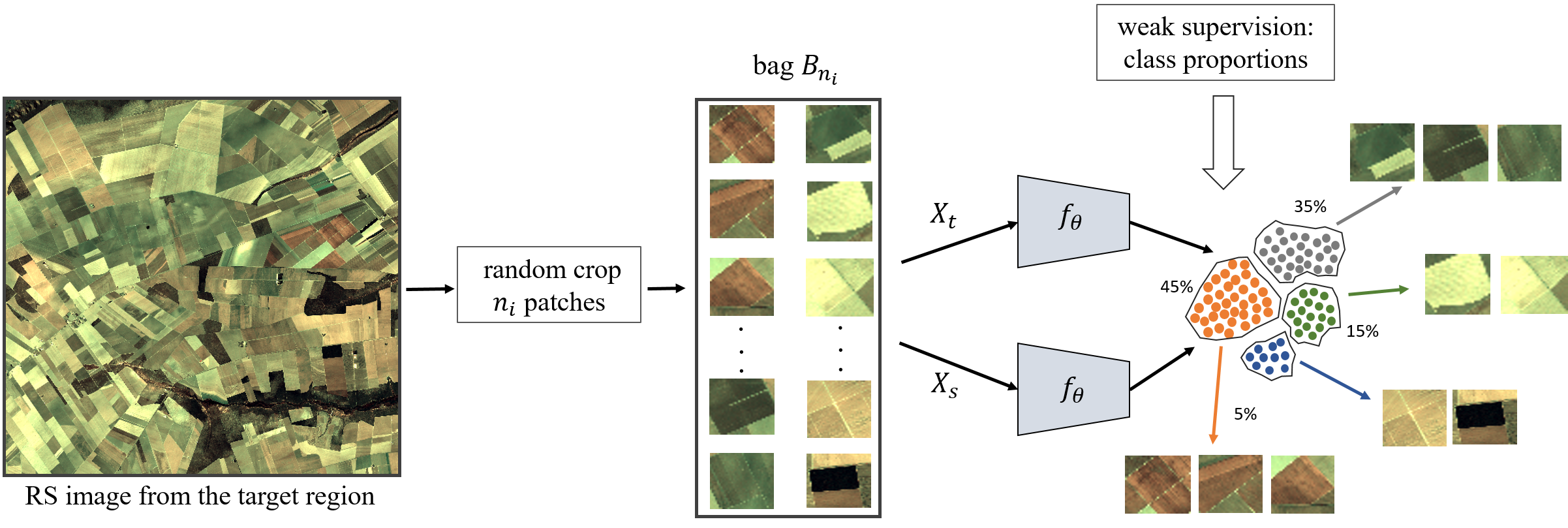}
\caption{General overview of LLP-Co for crop-type mapping. We first crop random image patches to form the input bag $B_i$ of image patches given a large-scale RS image. We then feed the bag of images to an augmentation pool to generate two transformed views of each image patch. Then, we forward the full batch of image views into a network that assigns them to various clusters constrained to their \emph{a priori} proportions. The colored arrows point to the image patches assigned to each cluster.}
\label{fig:framework}
\end{figure*}

Additionally, despite the unavailability of pixel-level annotations, many applications use approximate group-level labels like class proportions, as these are readily available from other sources, like a government census. For example, the National Agricultural Statistics Service of the United States Department of Agriculture\footnote{\url{https://www.nass.usda.gov/}} publishes annual statistics about US agriculture, such as area planted and harvested, yield, and production. The Brazilian Institute of Geography and Statistics (IBGE) collects statistical information about the population, agricultural farms, forests, and aquaculture of all municipalities in Brazil.\footnote{\url{https://www.ibge.gov.br/}} Similar census information is freely available for countries such as the United Kingdom,\footnote{\url{https://www.forestresearch.gov.uk/tools-and-resources/statistics/forestry-statistics/}} and many European countries.\footnote{\url{https://ec.europa.eu/eurostat}} In this sense, efficient learning from group-level labels has excellent potential for impacting many real-life EO applications, like crop-type identification. Previous studies have highlighted the possibility of using crop proportions to validate the generalization of trained networks using data from agriculture censuses \cite{de2020generalization}. Other researchers have gone further by artificially generating the native class proportions in the training set, which increases the accuracy by up to 30\% for binary cropland classification \cite{waldner2017impact}. 

Considering information about proportions, \emph{Learning from Label Proportions} (LLPs) is a weakly supervised classification approach that has gained popularity in the machine learning field in recent years \cite{quadrianto2009estimating,yu2013proptosvm,yu2014learning,qi2016learning,dulac2019deep,shi2020deep,scott2020learning}. LLP approaches aim to learn a sample-level classifier having only as reference the label proportions. In the traditional setting, it consists of dividing the available training samples into groups of bags where the label proportions are known for each bag. The objective function constrains the optimization process to finding the possible solution that accurately matches the proportions in each bag. Current state-of-the-art LLP methods are based on convolutional neural networks (DLLP \cite{dulac2019deep}), generative models (LLP-GAN \cite{liu2019learning}), optimal transport theory (LLP-GAN-PLOT \cite{liu2021two}), and contrastive learning \cite{llpl2021}. In this work, we embed priors about class proportions into a contrastive-learning framework for LLP. Specifically, the so-called \emph{Learning from Label Proportions with Prototypical Contrastive Clustering} (LLP-Co) approach \cite{llpl2021} relaxes the equipartition constraint in SwAV by incorporating the exact cluster proportions into the optimal transport solver. Here, the features and the cluster prototypes are learned online and end to end, making the method scalable to large datasets and thus, suitable for EO applications.

Among the existing EO applications, the accurate estimation of crop area extents and the crop-type distribution is indispensable for food security due to the increasing demand for food \cite{ramankutty2018trends,bodirsky2020ongoing}. Notwithstanding the success of unsupervised and supervised DL approaches that can integrate the spatial, spectral, and temporal contexts \cite{firat2014representation,romero2016unsupervised,Kussul2017,russwurm2018multi,ndikumana2018deep,zhong2019deep}, automatic crop mapping from RS images is still a challenging task. 

This paper proposes an online self-supervised contrastive-clustering method that disregards sample-level labeling but instead uses crop proportions from agricultural census data as priors to classify crops in regions with known expected crop proportions automatically. The general framework is based on the LLP-Co approach and is depicted in Fig.~\ref{fig:framework}. Thus, we address four specific research questions:

\begin{itemize}
    \item Is \textit{a priori} information about the exact class proportions sufficient for training a classifier for crop-type identification?
    \item Can a model converge to the real class proportion observed in the region using approximate class proportions?
    \item How does the bag size affect the final classification performance?
    \item Does the trained model generalize satisfactorily for different agricultural regions and modalities?
\end{itemize}

These four questions are extensively investigated in this paper for two large public crop-type datasets from two different agricultural regions in Brazil using Landsat-8 and Sentinel-1 images. We captured crop-proportion information in our experiments from public government census data for both regions.

\section{Related Works}
\subsubsection{Learning from label proportions} 

LLP aims to train a sample-level classifier using only information about bag proportions. Current state-of-the-art methods in this field use deep neural networks that enforce the probability outputs to match the bag proportion. In \cite{dulac2019deep}, the authors proposed a modified Kullback--Leibler (KL) divergence loss function and a function based on optimal transport with entropic regularization (ROT) to train a deep LLP model. Liu et al. \cite{liu2019learning} introduced LLP-GAN, which incorporates adversarial learning and is significantly better than other approaches. Recently, \cite{liu2021two} proposed a cyclic two-stage methodology that employs previous LLP models such as LLP-GAN in combination with a supervised learning stage that uses pseudo-labels generated by ROT as a reference. However, despite these recent advances, the models mentioned above employ the KL divergence between prior and posterior bag class proportions, and training a classifier solely with this constraint can lead to many valid hypotheses that match the prior distribution but still do not generalize well to unseen data. 

\subsection{Contrastive Learning and Clustering} 
Combining clustering and self-supervised learning has become a prominent feature representation learning approach for many computer-vision benchmarks \cite{asano2019self,caron2020unsupervised,li2020prototypical,li2020self}. Many of these approaches use prototypes to support clustering \cite{asano2019self}, in which each prototype defines a cluster centroid. In \cite{asano2019self}, the output features of a neural network serve as input to a clustering algorithm using a ROT solver based on the Sinkhorn--Knopp algorithm \cite{cuturi2013sinkhorn}, which assigns a prototype and creates ``pseudo-labels" for training the model. This approach alternates between cluster assignment, in which images from the entire dataset are clustered according to their features, and training, in which the prototype and code assignments are predicted from the different images. Such a procedure is usually described as offline clustering and requires multiple passes over the entire dataset \cite{caron2020unsupervised}. Recently, SwAV \cite{caron2020unsupervised} has combined contrastive learning and clustering in an end-to-end framework, achieving impressive results in self-supervised representation learning for several downstream computer-vision tasks, such as classification and object detection. Unlike \cite{asano2019self}, the method performs online clustering, since it computes the ``codes" (i.e., cluster assignments) using only the image features within a batch. 

\subsection{Self-supervised Methods for RS} 
Several self-supervised methods have been employed with success in EO applications \cite{ayush2021geography,li2021semantic,stojnic2021self,cai2021large,yue2021self}. In \cite{ayush2021geography}, the authors proposed a contrastive-learning approach that exploits the spatio-temporal information in satellite data and geolocation information, reporting impressive results in image classification, object detection, and semantic segmentation for RS datasets. Similarly, \cite{manas2021seasonal} proposed \emph{Seasonal Contrast} (SeCo), a self-supervised approach that combines contrastive learning with temporal and geolocation information to learn feature representation better for RS applications. Recently, a self-supervised learning method with adaptive distillation was proposed for hyperspectral image classification \cite{yue2021self}. \cite{cai2021large} proposed a contrastive-learning algorithm for large-scale hyperspectral image clustering named \emph{Spectral-Spatial Contrastive Clustering} (SSCC), which employs a twin neural network to conduct dual contrastive learning using a spectral-spatial pool of transformations. The contrastive objective function encourages within-cluster similarity while reducing between-cluster redundancy. Like other online clustering algorithms, the approach is trained end to end using batch-wise optimization. However, like many other contrastive-learning approaches, the network needs positive and negative sample pairs, making it computationally intensive.

\subsection{Contrastive Clustering and LLP} 
LLP-Co \cite{llpl2021} embeds cluster proportion priors into the SwAV approach to solve the LLP problem, filling the gap between supervised and weakly supervised approaches. Unlike its predecessor \cite{liu2021two}, the method consists of a one-stage end-to-end training scheme and does not employ KL divergence. It incorporates the exact cluster proportions into the ROT module, and the training goal is to find the network parameters that best describe the distribution of the overall training set. With this modification, the method goes beyond representation learning and inherently performs sample-level classification using the cluster assignments. Compared to previous approaches, LLP-Co learns the clusters jointly with the latent visual representation. In addition to the constraint on proportions, it implements a contrasting loss to force similar samples to gather together, which seems to reduce the solution space considerably.

\section{Methodology}\label{sec:methodology}
This section presents our methodology and the core concepts of the LLP-Co method \cite{llpl2021}. 
In LLP, the training samples are split into bags with known class proportions. These samples are used to train a sample-level classifier with a solver of choice. As in previous LLP approaches, the training dataset is composed of $N$ disjoint bags, where $B_i$ is the $i$th bag, which consists of a set of $n_i$ randomly selected samples $\mathcal{B}_i = \{(\mathbf{x}_{i,j})\}_{j=1}^{n_{i}}$, where $\mathbf{x}_{i,j}$ is the sample $j$ within the bag $i$. In this formulation, the training set can be expressed as $\mathcal{D} = \{(\mathcal{B}_i,\mathbf{w}_i)\}_{i=1}^N$, where $\mathcal{B}_i \cap \mathcal{B}_j = \emptyset$, $\forall i \neq j$. In an LLP multi-class problem with $K$ classes, $\mathbf{w}_i \in \Delta_K$ is the vector of label proportions for the $i$th bag such that (s.t.) $\sum_{k=1}^{K} \mathbf{w}_i^k = 1$, where the $\mathbf{w}_i^k$ element is the proportion of samples that belong to class $k$. In deep LLP approaches, a neural network commonly acts as a feature extractor. This is followed by a classification layer that delivers the probabilities vector $\mathbf{\tilde{p}}_{i,j} = p_\theta(\mathbf{y}|\mathbf{x}_{i,j})$ using a softmax operator, where $\theta$ represents the network parameters \cite{liu2019learning}. From here, we can estimate the bag-level label proportion as the summation of the element-wise posterior probability:
\[
\mathbf{\hat{w}}_i = \frac{1}{n_i}\sum_{j=1}^{n_i}\mathbf{\tilde{p}}_{i,j}
\]
and calculate the bag-level loss function as a standard cross-entropy loss function
\begin{equation}\label{lossllp}
  L(\hat{w},w) = -\frac{1}{N}\sum_{i=1}^{N}\mathbf{w}_i \log \mathbf{\hat{w}}_i.
\end{equation}

Following \cite{asano2019self,liu2021two}, and \cite{llpl2021}, Eq.~\eqref{lossllp} is then reformulated by encoding the label proportions as a posterior distribution:
\begin{equation}\label{losspq}
  L(p,q) = -\frac{1}{N}\sum_{i=1}^{N}\sum_{j=1}^{n_{i}}\sum_{k=1}^{K} \frac{q(y^k|\mathbf{x}_{i,j})}{n_i} \log p_\theta(y^k|\mathbf{x}_{i,j}).
\end{equation}
The final LLP function objective is then:
\begin{equation}\label{learobj}
\min_{(p,q)} L(q,p), \quad \text{s.t.} \quad \forall y: q(y^k|\cdot) \in [0,1]
\end{equation}
\begin{equation}\label{learobj1} \sum_{j=1}^{n_{i}}q(y^k|\mathbf{x}_{i,j}) = \mathbf{w}_i^k n_i,
\end{equation}
where the proportion constraint ensures that each label $k$ contains overall $\mathbf{w}_i^k n_i$ samples.

As discussed in \cite{asano2019self,liu2021two}, and \cite{llpl2021}, this equation is an instance of the ROT problem and can be solved efficiently using the Sinkhorn--Knopp algorithm. Let 
\[
\mathbf{P}_{i,j}^y = p_\theta(y|\mathbf{x}_{i,j})\frac{1}{n_i}
\]
be the $K\times n_i$ joint probabilities matrix estimated by the model and 
\[
\mathbf{Q}_{i,j}^y = q(y|\mathbf{x}_{i,j})\frac{1}{n_i}
\]
be the $K\times n_i$ matrix of assigned joint probabilities for bag $\mathcal{B}_i$. In the LLP approach, $\mathbf{Q}_i$ is expected to split the data within the bag by respecting prior information about label proportions. Hence, we add the constraint to the transportation polytope and rewrite the objective function in Eq.~\eqref{learobj} for the $i$th bag as an OT solver and add the entropic regularization term. Thus, the modified objective function becomes
\begin{equation}\label{finalform}
    \min_{\mathbf{Q}_i\in U(\mathbf{w},\mathbf{a})_i} \langle \mathbf{Q}_i, -\log \mathbf{P}_i\rangle + \varepsilon h(\mathbf{Q}_i),
\end{equation}
where
\begin{equation}\label{otconstrain}
    U(\mathbf{w},\mathbf{a})_i := \{\mathbf{Q}_i \in \mathbb{R}_+^{K\times n_i} | \mathbf{Q}_i\mathbf{1}_{n_{i}} = \mathbf{w}_i, \mathbf{Q}_i^T\mathbf{1}_K = \mathbf{a}\}.
\end{equation}

Here $U(\mathbf{w},\mathbf{a})_i$ is the matrix space of possible solutions for bag $\mathcal{B}_i$, $\mathbf{w}$ is the class proportions vector, and $\mathbf{a} = (1/n_i)\mathbf{1}_{n_{i}}$ is a normalizing constraint that reinforces that samples are assigned to a single cluster \cite{genevay2019differentiable}. All matrices in $U$ are non-negative such that $\mathbf{w}$ and $\mathbf{a}$ are their row and column marginals, respectively. Using DL nomenclature, the bag size of LLP is equivalent to the batch size and the optimization process is performed using gradient descent for each batch.

\subsection{Learning from Label Proportions with Prototypical Contrastive Clustering} 
LLP-Co is an online self-supervised clustering method that employs a convolutional neural network to learn a latent visual representation that delivers consistent cluster assignments between views or augmentations of the same input image. The method uses a contrastive loss by comparing image cluster assignments instead of their features. Given two or more augmented views of the same input image, the online clustering algorithm uses one augmented view to compute soft targets using an OT solver and the other augmented views to predict these targets using the cross-entropy loss function. Unlike SwAV \cite{caron2020unsupervised}, LLP-Co replaces in the ROT solver the equipartition constraint with cluster size constraints using Eq.~\eqref{finalform}.

In LLP-Co, each image $j$ within a bag $\mathcal{B}_i$ of images is transformed into two augmented views, which are fed to an encoder network to extract the two corresponding sets of features $\mathbf{z}_{i,j}^s,\mathbf{z}_{i,j}^t \in \mathbb{R}^{m}$. The features are then projected to the unit sphere and mapped to one of the $K$ trainable prototypes $\mathbf{V} = [\mathbf{v}_1,\dots,\mathbf{v}_k]$ to derive the code assignments for each of the views $\mathbf{c}_{i,j}^s$ and $\mathbf{c}_{i,j}^t$. The contrastive loss function performs a ``swapped" procedure that predicts the assignment of one feature from the code of the other. Hence, the feature extractor network and prototype weights are jointly trained to minimize the subsequent loss for all samples $j$ within bag~$i$:
\begin{equation}\label{loss2v}
    L_{swap}(\mathbf{z}_{i,j}^s,\mathbf{z}_{i,j}^t) = \ell (\mathbf{z}_{i,j}^s,\mathbf{c}_{i,j}^t) + \ell (\mathbf{z}_{i,j}^t,\mathbf{c}_{i,j}^s),
\end{equation}
where each term represents the cross-entropy loss between the code and the probability obtained as the softmax function of the dot product between the features and all the prototypes:
\begin{equation}
    \ell(\mathbf{z}_{i,j}^t,\mathbf{c}_{i,j}^s) = -\sum_{k} \mathbf{c}_{i,j}^{s(k)} \log \mathbf{p}_{i,j}^{t(k)}, 
\end{equation}
where
\begin{equation}
\mathbf{p}_{i,j}^{t(k)} = \frac{\exp ((\mathbf{z}_{i,j}^t)^\mathsf{T}\mathbf{v}_k/\tau)}{\sum_{k'} \exp ((\mathbf{z}_{i,j}^t)^\mathsf{T}\mathbf{v}_{k'}/\tau)}.
\end{equation}
The intuition behind the swap loss function is that if $\mathbf{z}_{i,j}^t$ and $\mathbf{z}_{i,j}^s$ contain similar information, it should be possible to predict the code $c_s$ (i.e., soft class) from the feature $\mathbf{z}_{i,j}^t$, and vice versa. In other words, if the features share similar semantics, their target codes will be similar.

To compute the codes, LLP-Co uses the entropic regularized OT as formulated in~\eqref{finalform}, which ensures the samples in the bag are partitioned according to the bag-level cluster-size proportions. For the $i$th bag, let $\mathbf{Z}_i$ be the feature vectors, $\mathbf{V}$ the prototypes, and  $\mathbf{Q}_i$ the codes that perform the transportation subject to the proportion constraints. Using the notation of \cite{caron2020unsupervised} and \cite{llpl2021}, the objective is to optimize $\mathbf{Q}_i$ to maximize the similarity between $\mathbf{Z}_i$ and $\mathbf{V}$ as follows:
\begin{equation}\label{ot}
    \mathbf{Q}_{i}^{*} = \max_{\mathbf{Q}_i\in U} \mathbf{Tr} (\mathbf{Q}_i^\mathsf{T} \mathbf{V}^\mathsf{T} \mathbf{Z}_i) + \varepsilon h(\mathbf{Q}_i),
\end{equation}
where $h(\mathbf{Q}_i)$ is the entropy function and $\varepsilon$ is a trade-off parameter that controls the smoothness of the prediction. In practice, a high $\varepsilon$ implies strong entropy regularization, which can lead to the well-known problem of model collapse, in which all samples are mapped to a unique representation; thus, $\varepsilon$ is generally set to a low value. 

In Eq.~\eqref{ot}, $\mathbf{V}^\mathsf{T} \mathbf{Z}_i$ is a $K\times n_i$ matrix that represents the cosine similarities between all given samples and each prototype within the bag, and $\mathbf{Q}$ are codes that perform the transportation, i.e., they weight the similarities, restricted to the proportion constraint. This formulation is equivalent to the learning objective in Eq.~\eqref{finalform}. Using the regularization term allows us to write the optimization problem as a normalized exponential matrix, which can easily be computed through iterative matrix multiplication using the Sinkhorn-Knopp algorithm \cite{cuturi2013sinkhorn}. The Sinkhorn-Knopp algorithm iteratively optimizes $\mathbf{Q}_i\in U$ to maximize the similarity between sample features and their respective assigned clusters while keeping the entropy relatively high. We outline the learning procedure for two views in Algorithm~\ref{alg:algorithm}. For more information about the LLP-Co method, see \cite{llpl2021}.

\begin{algorithm}[tb]
\caption{LLP-Co training loop using two views}
\label{alg:algorithm}
\textbf{Input}: Samples and proportions of bags $\mathcal{D} = \{(\mathcal{B}_i,\mathbf{w}_i)\}_{i=1}^N$\\ \textbf{Input}: Epochs $\varepsilon > 0$ \\
\textbf{Initialize}: Encoder network $f_\theta$ and prototypes $\mathbf{V}$ with random weights
\begin{algorithmic}[1] 
\FOR{i = 1 to epochs}
\FOR{each $\mathcal{B}_i$ in $\mathcal{D}$}
\STATE Generate two random views $\mathbf{X}_{i}^{t,s}$\\
\STATE Obtain the feature vectors $\mathbf{Z}_{i}^{t,s}$\\
\STATE Compute the prototype scores $\mathbf{V}^\mathsf{T}\mathbf{Z}_{i}^{t,s}$\\
\STATE Compute the codes $\mathbf{Q}_{i}^{t,s}$ constrained to $\mathbf{w}_i$\\
\STATE Convert prototype scores to probabilities $\mathbf{P}_{i}^{t,s}$\\
\STATE Compute the loss using $L_{swap}$\\
\STATE Update $\theta$ and $\mathbf{V}$ with a gradient step\\
\ENDFOR 
\ENDFOR
\end{algorithmic}
\end{algorithm}

\subsection{Learning from Global Proportions with LLP-Co}
In this work, we propose to use LLP-Co in a more realistic scenario, in which we do not have access to the real bag label proportions to train the network. Instead, we have access only to the global class proportions. For this, the training samples are also split into $N$ disjoint bags, where $B_i$ is the $i$-th bag, which consists of a set of $n_i$ randomly selected samples. Unlike traditional LLP training, the training set is $\mathcal{D} = \{(\mathcal{B}_i,\mathbf{w})\}_{i=1}^N$, where $\mathbf{w} \in \Delta_K$ is the vector of global label proportions, which is the same for all bags $B_i$ and s.t.\ $\sum_{k=1}^{K} \mathbf{w}^k = 1$, where the element $\mathbf{w}^k$ is the proportion of samples that belong to class $k$.

\section{Experimental Design}\label{sec:experiments}
\subsection{Crop-type Datasets}
\subsubsection{Study Areas} 
\begin{figure}[th!]
\centering
\includegraphics[scale=0.26]{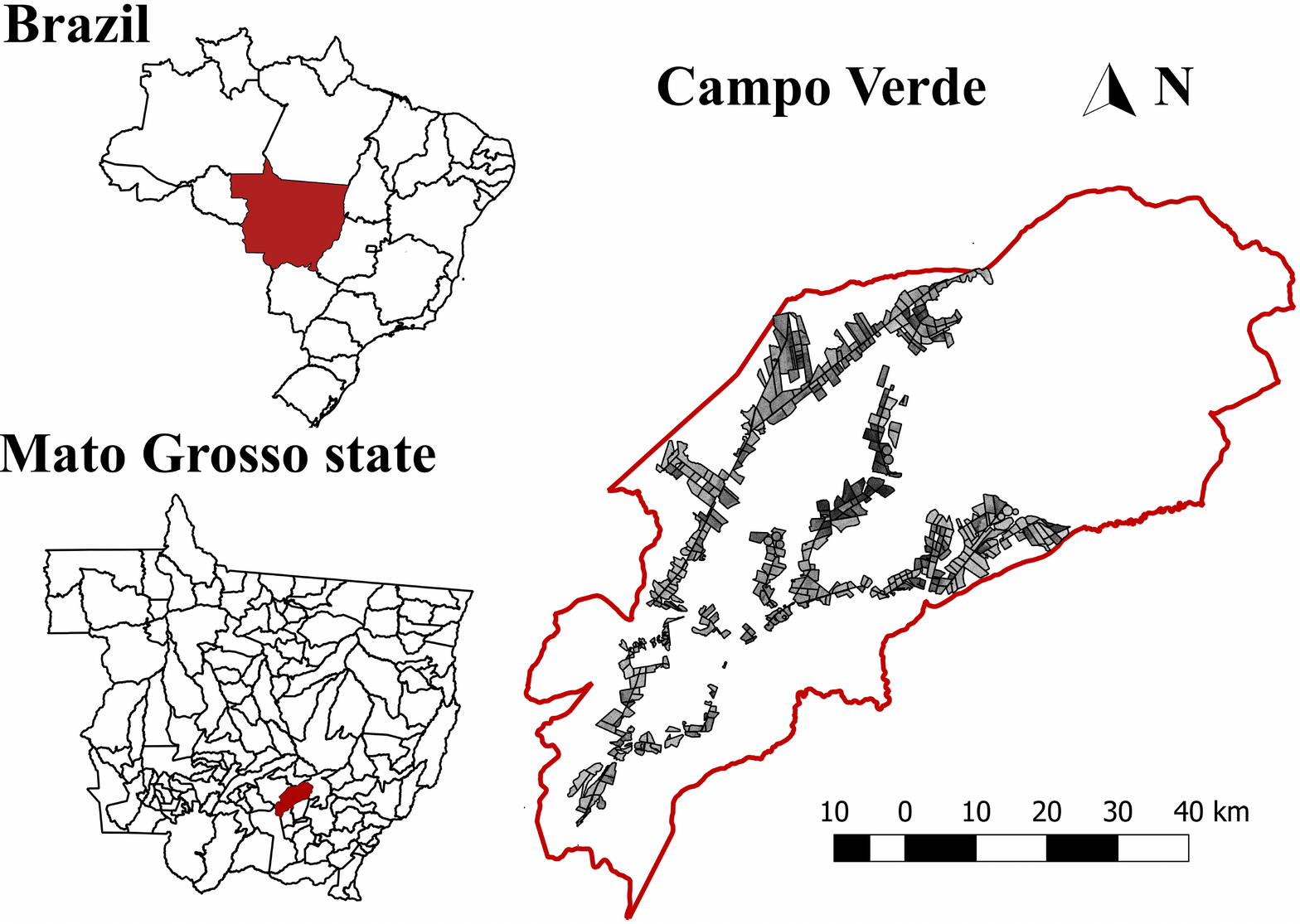}
\caption{Location of images in CV.}
\label{fig:cv_map}
\end{figure}

\begin{figure*}[bt!]
\centering
\includegraphics[scale=0.23]{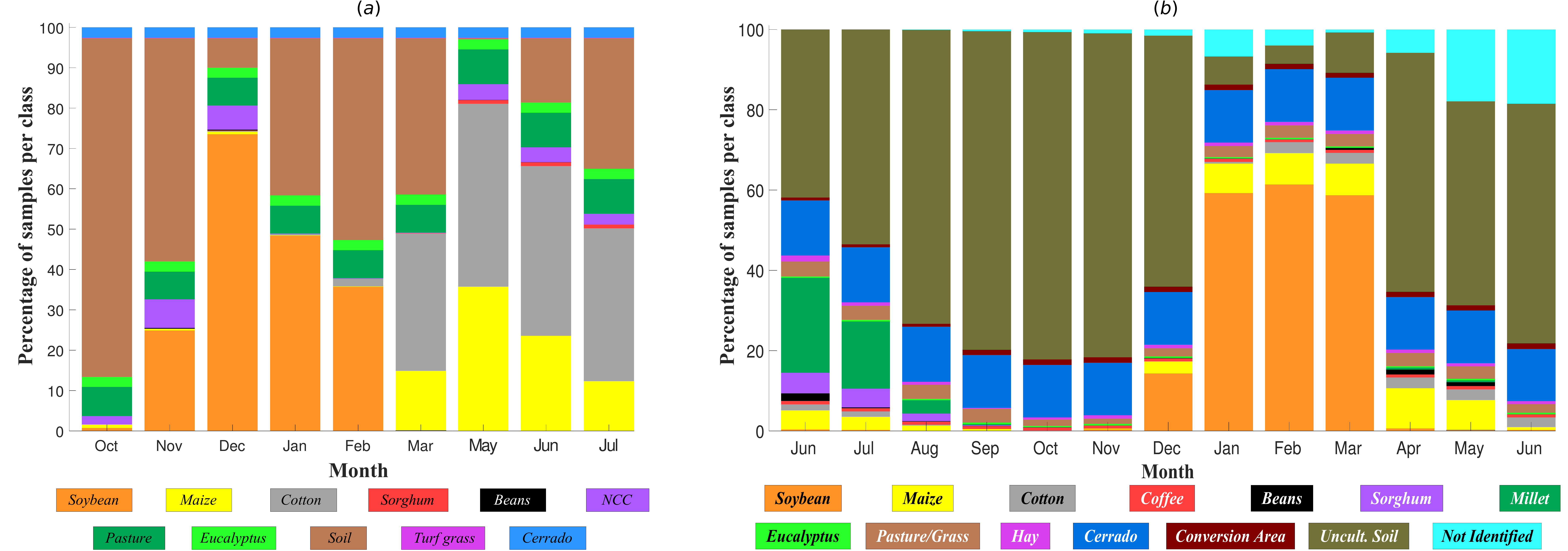}
\caption{Class distributions for the CV (left) and LEM datasets (right). NCC refers to \textit{Non-Commercial Crops}.}
\label{fig:class_distr}
\end{figure*}

The experiments were conducted with two datasets from two agricultural regions in Brazil, both publicly available. The first region of interest is in Campo Verde (CV) municipality, Mato Grosso, at a latitude of $15\degree32'48"$ south and a longitude of $55\degree10'08"$ west (Fig.~\ref{fig:cv_map}). It has an extension of 4782~km$^2$ \cite{sanches2018campo}. The dataset contains preprocessed Sentinel-1A and Landsat-8 images and 513 reference polygons, corresponding to 6 million pixels. The images are for the period between October 2015 and July 2016, so that the class distribution varies greatly over time (Fig.~\ref{fig:class_distr}a). \textit{Soybean} was the main crop type grown from October 2015 to February 2016. It was replaced by \textit{cotton} and \textit{maize} in the following months.\footnote{The CV database is available from IEEE Dataport at \url{https://ieee-dataport.org/documents/campo-verde-database}.} 

\begin{figure}[th!]
\centering
\includegraphics[scale=0.45]{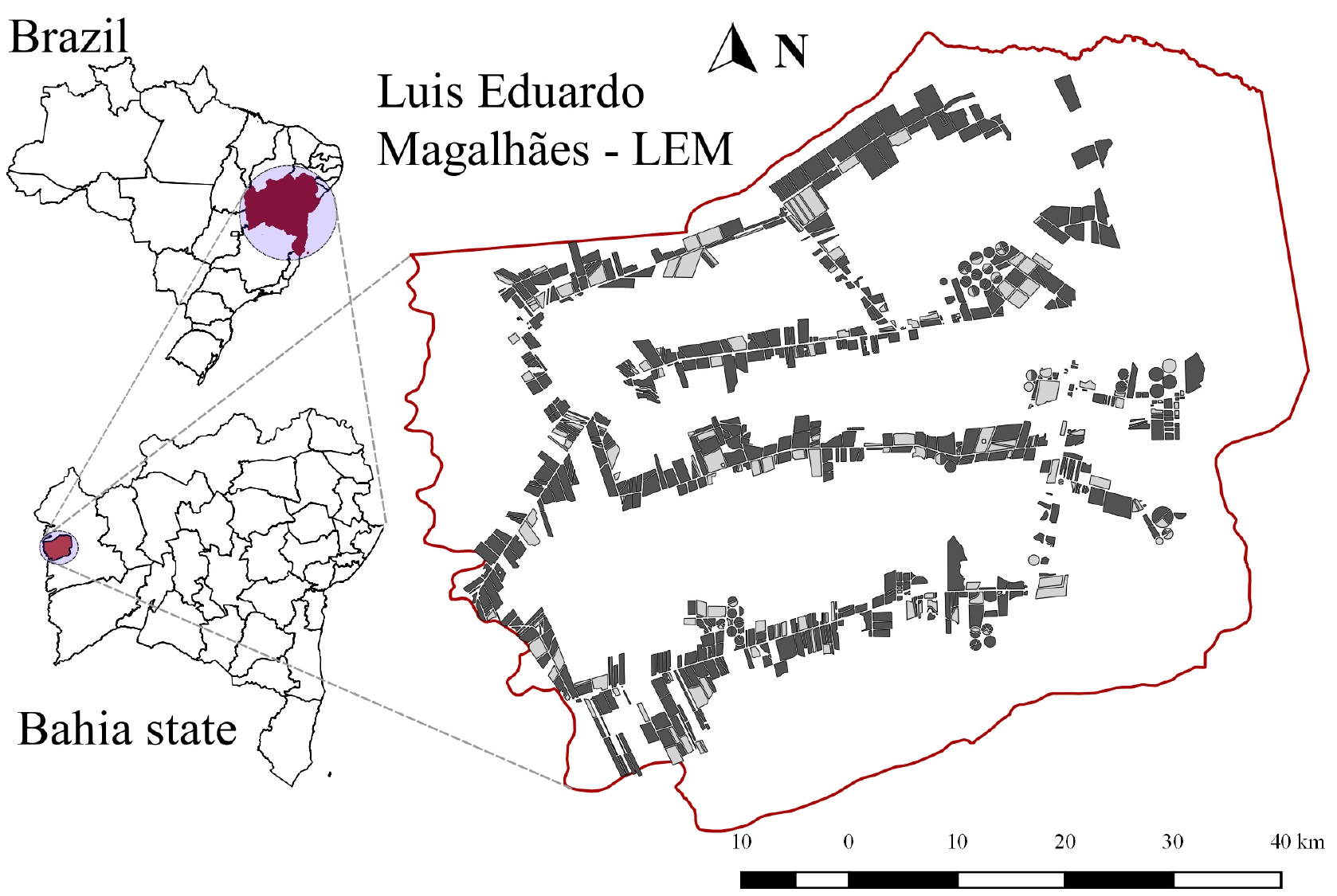}
\caption{Location of images in LEM.}
\label{fig:lm_map}
\end{figure}

The second test site is in Luis Eduardo Magalh\~{a}es (LEM) municipality, Bahia state. It has an area of 3940~km$^2$ and is at a latitude of $12\degree05'31"$ south and longitude $45\degree48'18"$ west (Fig.~\ref{fig:lm_map}) \cite{sanches2018lem}. It contains 807 reference polygons and consists of preprocessed Sentinel-1, Sentinel-2, and Landsat-8 images, obtained between June 2017 and June 2018. Like CV, the class distribution in the LEM dataset was not uniform over the year, as shown in Fig.~\ref{fig:class_distr}b. The main crop types were \textit{soybean}, \textit{maize}, \textit{cotton}, and \textit{millet}.\footnote{The LEM database is freely accessible at \url{http://www.dpi.inpe.br/agricultural-database/lem/.}} 

\subsubsection{Class Proportion Availability}
For our experiments, we used publicly available census data from the IBGE. The dataset contains the proportions of cultivated areas in both municipalities. The institute's website has public reports with crop area estimates for different agricultural regions in Brazil per year, including municipality-level information. The institute's method for crop area estimation includes interviews and field visits and therefore, does not rely on field imagery. Table~\ref{tab:props} gives the percentages of the overall area planted with major crops in the regions as estimated by the IBGE, as well as the percentages observed in the annotated datasets.

\begin{table}[th]
\caption{Planted area (\%) for each crop in CV and LEM.}
\label{tab:props}
\centering
\begin{tabular}{ccccccc}
\toprule
\multirow{2}{*}{Dataset} & \multicolumn{3}{c}{IBGE} & \multicolumn{3}{c}{Annotated data}\\
\cmidrule(lr){2-4} \cmidrule(lr){5-7}
& Cotton & Maize & Soybean & Cotton & Maize & Soybean\\\midrule
CV & 20.05 & 22.32 & 54.36 & 45.30 & 35.8  & 73.20 \\ 
LEM & 4.95 & 7.06 & 78.63 & 2.70 & 7.80 & 61.41 \\\bottomrule
\end{tabular}
\end{table}

\subsection{Experimental protocol}
\begin{figure*}[t]
    \centering
		\begin{subfigure}[b]{0.99\linewidth}
             \centering
             \includegraphics[width=0.99\textwidth]{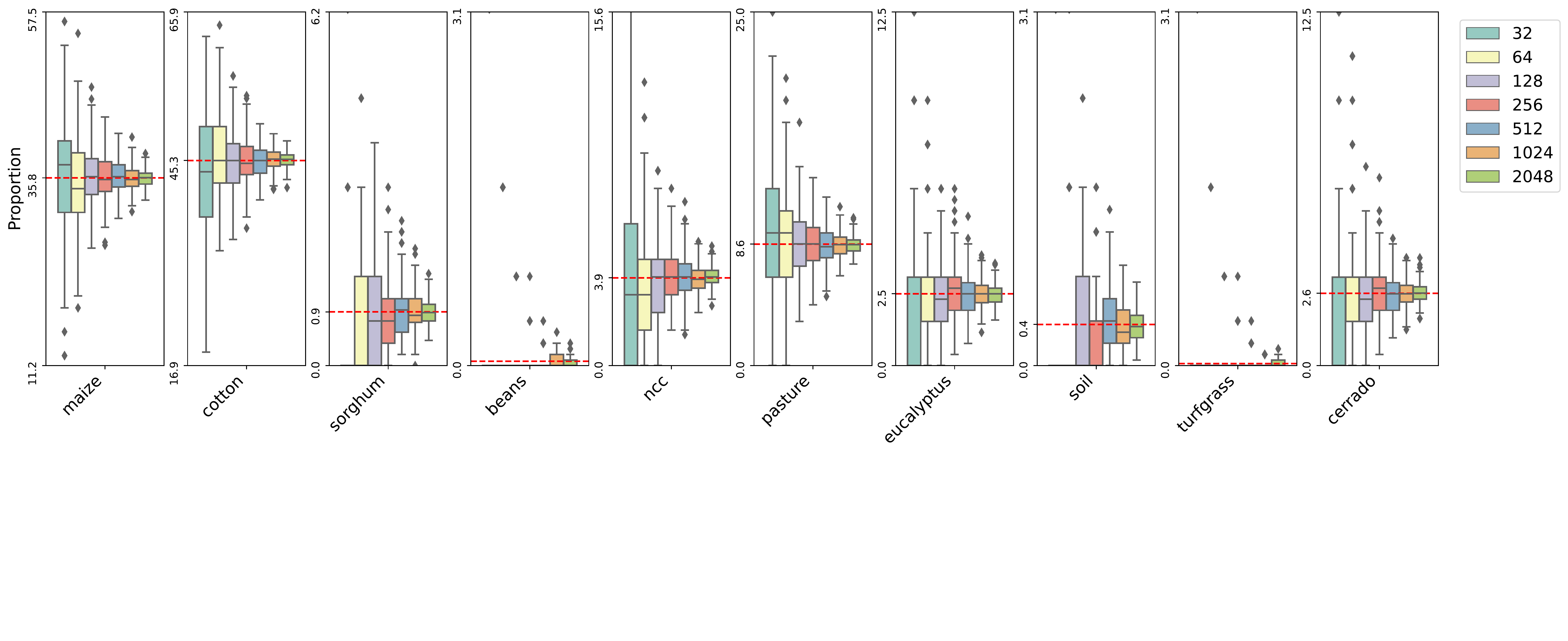}
    \end{subfigure}
 		\begin{subfigure}[b]{0.99\linewidth}
              \centering
              \includegraphics[width=0.99\textwidth]{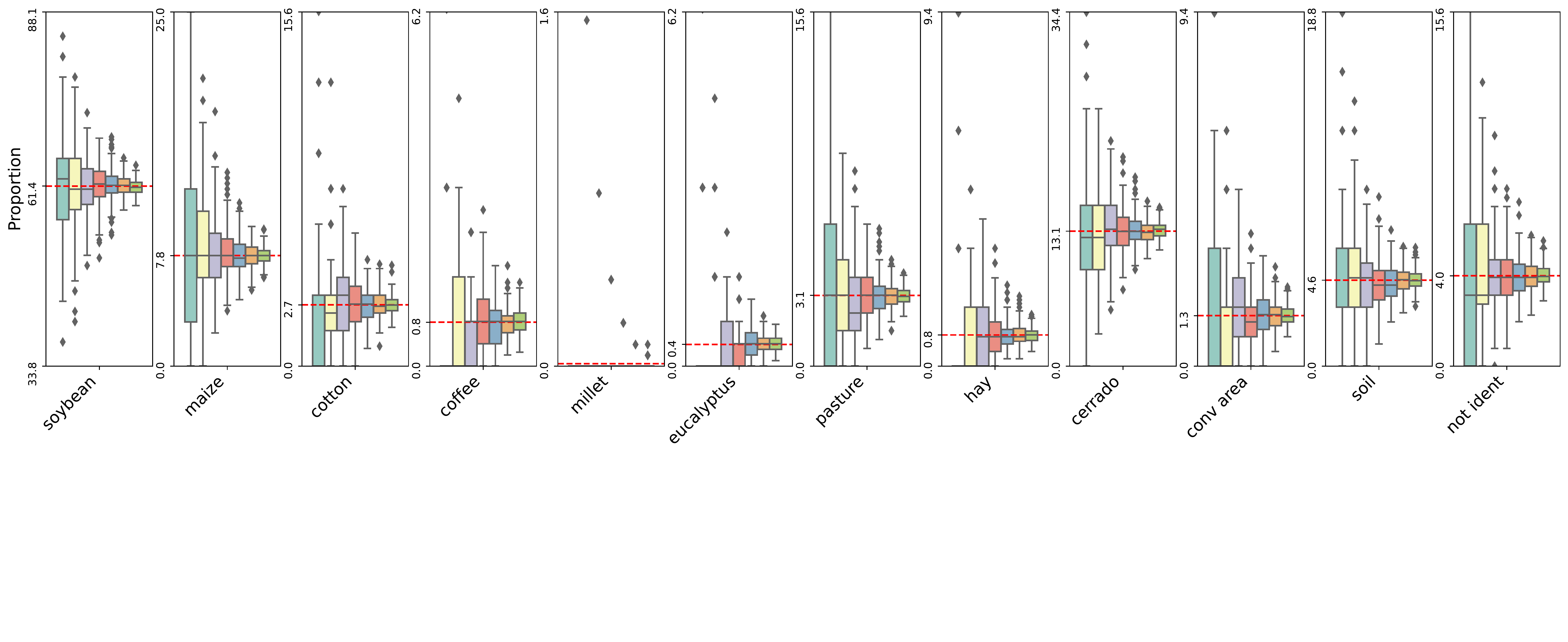}
     \end{subfigure}
	\caption{Box plots of class proportions for different bag sizes for 300 iterations for CV (top) and LEM (bottom). The red dashed lines are the class proportions observed in the datasets for each crop type. \textit{Soybean}, \textit{maize}, and \textit{cotton} are the major crop types.}
	\label{fig:boxplot}
\end{figure*}

Our experiments focused on the first Brazilian crop harvest for each region, which was in the wet season. To assess the applicability of the methodology to different data sources, we employed both optical and synthetic-aperture radar (SAR) data. For the CV dataset, we considered the only cloud-free optical image available for this period, i.e., one taken in May 2016. For the LEM dataset, we used all the multi-temporal SAR images captured between December and April, taking as reference the labels for February 2017, which can be considered the peak of the season. Before training, we upsampled the optical data to match the corresponding SAR image resolution and split the dataset so that for CV, 50\% of the polygons were used for training and 50\% for testing, and for LEM, 75\%  were used for training and 25\% for testing.

Fig.~\ref{fig:boxplot} shows the class proportions for each crop type in the CV and LEM annotated datasets for 300 iterations for different bag sizes. As the bag size increases, the proportions converge to the native class proportions observed in the dataset (red dashed lines in Fig.~\ref{fig:boxplot}). Nonetheless, note the significant differences in the class proportions between major and minor crops. We hypothesize that for highly unbalanced datasets, our approach will find the set of parameters that delivers, on average, the best performance. Without any extra signal besides the global proportions, the model will be biased to maximize the performance for the majority classes. For this reason, this study focused primarily on the classification of the major crop types. 

We evaluated the LLP-Co method under four scenarios with different training parameters. The first two scenarios were the main contribution of this paper, since they focus on the use of global class proportions to identify the major crop types grown in the study sites. For the first scenario, we used the global class proportions from the annotated dataset, and for the second scenario, we used the class proportions estimated for the census data. Unlike the training schemes commonly used in LLP approaches, which calculate the exact class proportion for each bag in a supervised fashion, these scenarios used only weak information in the form of global class proportions. For scenarios three and four, we evaluated the LLP-Co method for all crop types grown in the study sites, which allowed us to assess the sensitivity of the model to highly unbalanced class proportions under the traditional LLP training setting and using only global information. A detailed description of each scenario is given below. Table~\ref{tab:variants} summarizes the proposed experiments.

\subsubsection{Scenario one (SI)} In this scenario, we used as prior proportions the global proportions in the datasets for only the major crops in the target regions for the date of interest. The remaining crop types were grouped into class \textit{others}. For a given bag size $n_i$, we created the training bag $\mathcal{B}_i$ by randomly selecting $n_i$ image patches from the annotated region, so that each sample within the bag was unique. Considering the native class proportions observed in the datasets (red dashed lines in Fig.~\ref{fig:boxplot}), the global proportions for CV were 35.8\% for class \textit{maize}, 45.3\% for class \textit{cotton}, and 18.9\% for class \textit{others}. For the LEM dataset, the proportions were 61.4\% for class \textit{soybean} and 38.6\% for class \textit{others}. To generate the image patches, we randomly cropped patches from the whole dataset (i.e., the annotated area) and used the class of the central pixel of the patch. Since we used global class proportions and considered the mean bag proportions for the different bag sizes presented in Fig.~\ref{fig:boxplot}, we adopted a large bag size, i.e. $n_i = 2048$, for this scenario.  

\subsubsection{Scenario two (SII)} For the second scenario, which is the most relevant in our analysis, we defined a training schema that uses only the crop proportions available from a census or another source, in this case, the IBGE (Table~\ref{tab:props}). Since we do not have labels for the whole municipality, it was necessary to discard the non-agricultural regions (e.g., urban areas, forests, and rivers). Following \cite{9026754}, we masked the non-agricultural area using the standard deviation of the normalized difference vegetation index (NDVI) from the available optical images and removed pixels with a standard deviation of less than 25\%. Then, we created the training bags $\mathcal{B}_i$ by randomly selecting $n_i$ image patches from the whole crop area (which also included non-annotated regions). We used as priors the crop proportions for the whole municipality. To generate the image patches, we randomly cropped patches from the whole crop area within the municipality boundaries and used the class of the central pixel of the patch. Since we used estimated proportions and the means and standard deviations of the proportions in Fig.~\ref{fig:boxplot} for this scenario, we also defined an experiment with a large bag size, i.e., $2048$. Notice that the proportions reported in the census differ considerably from those observed in the annotated datasets. Thus, this scenario allowed us to evaluate whether approximated class proportions are enough for the model to converge to the actual class proportion.

\subsubsection{Scenario three (SIII)} This scenario adopted the training scheme commonly used in LLP approaches, where the exact class proportions for each crop type within each bag are known. For a given bag size $n_i$, we created the training bag $\mathcal{B}_i$ by randomly selecting $n_i$ image patches from the training set, so that each sample within the bag was unique, and calculated the exact class proportion for this bag. To generate the image patches, we randomly cropped patches from the annotated training region and used the class of the central pixel of the patch. Since we were working with exact class proportions for each bag, we could use small and large bag sizes. Hence, we defined six experiments with different bag sizes: $n_i = [32, 64, 128, 256, 512, 1024]$. This allowed us to assess the sensitivity of the model to this parameter. 

\subsubsection{Scenario four (SIV)} For scenario four, we defined a training scheme similar to scenario SI but considering the global proportion of the crops in all classes observed in the dataset as prior information. For a given bag size $n_i$, we created the training bag $\mathcal{B}_i$ by randomly selecting $n_i$ image patches from the annotated region, so that each sample within the bag was unique. For the prior proportions, we did not use the exact proportions within the bag but the global proportions of the dataset. Hence, to generate the image patches, we randomly cropped patches from the whole dataset, not only from the training region, and used the class of the central pixel of the patch. For this scenario, we defined four experiments with different bag sizes: $n_i = [256, 512, 1024, 2048]$. These are relatively large bag sizes, and we expected that the approach would perform better as the bag size increased.

\subsubsection{Baseline} As a baseline, we adopted the SwAV method with the equipartition constraint, i.e., no priors for the proportions. For a fair comparison, we used the same experimental setup as the above scenarios but without the proportions. We call these models SwAV1 (which corresponds to scenario SI), SwAV2 (scenario SII), and SwAV3 (scenario SIV). 

\begin{table*}[ht]
\caption{Proposed experiments.}
\label{tab:variants}
\centering
\begin{tabular}{lccc}
\toprule
\textbf{Experiment} & \textbf{Proportions} & \textbf{Training region} & \textbf{Classes}\\\midrule
Scenario I (SI) & Global proportions from the annotated data & Annotated region & Major crops and class \textit{others}\\
Scenario II (SII) & Global proportions from census data & Masked agricultural region & Major crops and class \textit{others}\\
Scenario III (SIII) & Exact bag proportion & Annotated data & All classes\\
Scenario IV (SIV) & Global proportions from the annotated data & Annotated region & All classes\\\midrule
Baseline SwAV1 & Equipartition constraint & Annotated region & Major crops and class \textit{others}\\
Baseline SwAV2 & Equipartition constraint & Masked agricultural region & Major crops and class \textit{others}\\
Baseline SwAV3 & Equipartition constraint & Annotated region & All classes\\\bottomrule
\end{tabular}
\end{table*}

\subsection{Implementation Details}\label{experimental}

\subsubsection{Architecture and Training} 
Our method uses a modified ResNet18 as the backbone architecture. We set the stride of the first convolution to 1. The ResNet18 is then followed by a projection head that projects the features to a $1024$-dimensional space. As in \cite{llpl2021}, all models were trained using stochastic gradient descent, with a weight decay of $10^{-6}$ and an initial learning rate of $0.1$. We warmed up the learning rate during five epochs and then used the cosine learning rate decay \cite{loshchilov2016sgdr} with a final value of $0.0001$. We followed \cite{caron2020unsupervised} and set the softmax temperature $\tau$ to 0.1. The prototypes were frozen during the first epoch. In scenario SIII, the models with small bag sizes converged faster. Hence, we trained these models for $200$ epochs and the model with the largest bag size for $400$ epochs. We used image patches of $21 \times 21$, and at each epoch for each dataset, we randomly selected 200,000 image patches to create the random bags. We augmented the datasets with random rotations, mirroring, and random resizing to obtain two views. 

\subsubsection{Hyperparameters for the Cluster Assignment using OT} 
The weight of the entropy term $\varepsilon$ was set to $0.05$, and the number of Sinkhorn iterations to 5. Following \cite{llpl2021}, we used a hard assignment for the bag size of 32 in scenario SIII and a soft assignment for the other bag sizes. For the baseline SwAV method, we employed only a soft assignment. The number of clusters for all models was set to the number of classes in each corresponding scenario and to 30 for the baseline SwAV method. The authors of the SwAV methodology recommend setting the number of prototypes to more or less than three times the number of classes; hence, we selected 30 for both datasets. For scenario SI and SII, we considered only the major crops listed in Table~\ref{tab:props}. We grouped the others into the class ``others." Hence, considering the dates we focused on for each dataset, we set the number of clusters as three (\textit{cotton}, \textit{maize}, and \textit{others}) and two (\textit{soybean} and \textit{others}) for CV and LEM, respectively.

\subsubsection{Evaluation metrics} For the four scenarios, we used the trained model to perform online cluster assignment and evaluated the quality of the clustering/classification using metrics such as cluster accuracy ($Acc_H$), normalized mutual information (NMI), and adjusted rand index (ARI). To compute these metrics, we employed a Hungarian match \cite{kuhn1955hungarian} between the true class labels and the cluster assignment. Since we use the proportion priors, we can also report the classification accuracy by considering the network's cluster assignment as the predicted label ($\mathsf{Acc}_\mathsf{P}$). That is, if a sample was assigned to the prototype $\mathbf{v}_1$, the corresponding labels were~$1$. To evaluate the baseline model, we used the embeddings $\mathbf{z}$ generated by the ResNet18 backbone and then performed $k$-means clustering. We also evaluated the models using the $k$-nearest-neighbor (kNN) classification \cite{wu2018unsupervised}. For a feature $\mathbf{z}$ in the test set, we took the top $25$ nearest neighbors from the training set and performed majority voting to assign the label.

\section{Results and Discussion}

\subsection{Results for SI and SII}
Table~\ref{tab:s1s2} shows the performance with the test set for scenarios SI and SII as well as for the SwAV baseline methods (SwAV1 and SwAV2), for both datasets. The model was trained on only the major crop types and class \textit{others}. We ran the $k$-means algorithm over the frozen features using five different seeds for the baseline methods and report the average values. Notice that the classification results vary by up to 30\% for the five runs depending on the seed. 

\begin{table*}[ht!]
    \centering
        \caption{Test performance for the CV and LEM datasets for the scenarios SI and SII and the baseline SwAV model.}
        \label{tab:s1s2}
        \begin{tabular}{ccccccccc}
        \toprule
        \multirow{2}{*}{Metric} & \multicolumn{4}{c}{CV} & \multicolumn{4}{c}{LEM}\\
        \cmidrule(lr){2-5} \cmidrule(lr){6-9}
        & SI & SII & SwAV1 & SwAV2 & SI & SII & SwAV1 & SwAV2 \\\midrule
        $Acc_P$ & 94.1 & 35.0 & -- & -- & 90.5 & 85.5 & -- & --\\
        $Acc_H$ & 94.1 & 80.5 & 74.4 & 74.7 & 90.5 & 85.5 & 84.1 & 53.5\\
        kNN & 92.0 & 91.0 & 89.2 & 89.7 & 98.8 & 95.3 & 96.0 & 98.3\\
        ARI & 0.83 & 0.48 & 0.50 & 0.41 & 0.66 & 0.49 & 0.46 & $-0.03$  \\
        NMI & 0.76 & 0.51 & 0.50 & 0.44 & 0.53 & 0.47 & 0.36 & 0.12\\
        \bottomrule
    \end{tabular}
\end{table*}

The classification accuracy using the Hungarian algorithm for scenario SI reported promising results, achieving accuracies of 94.1\% and 90.5\% ($Acc_P$ and $Acc_H$) for the CV and LEM datasets, respectively. The cluster quality metrics ARI and NMI show that performance was better for the CV dataset than for the LEM dataset. This may be due to the different data sources used in our experiments, since crop-type mapping from SAR imagery is more challenging than from optical data. 

For scenario SII (which uses IBGE proportions), the model reached an accuracy of 80.5\% and 85.5\% for the CV and LEM datasets, respectively, in terms of $Acc_H$. That represents a significant drop in performance of about 14 percentage points for the CV dataset compared to SI. In addition, unlike scenario SI, we observed cluster swapping for CV; therefore, $Acc_P < Acc_H$. This behavior was somewhat expected, since the estimated proportions for the area of interest according to the government census differ from the proportions found in the annotated dataset (Table~\ref{tab:props}). The IBGE proportion for class \textit{maize} is higher than the proportion for class \textit{cotton}, while the opposite can be observed for the annotated set. In contrast, for the LEM dataset, the estimated proportions according to the government census reflect the native proportions in the annotated dataset, with class \textit{soybean} being the major crop type. Thus, $Acc_P = Acc_H$. The performance measured by the ARI and NMI metrics was worse, with values ranging from 0.47 to 0.49. 

Comparing the results from LLP-Co with those from the baseline SwAV models, we observe that including class proportions in the learning process was always beneficial. Scenario SI performed better than the baseline SwAV1 method by 20 and 6 percentage points for the CV and LEM datasets, respectively. A more significant improvement was observed for scenario SII with the LEM dataset, reaching an enhancement of 32 percentage points compared with SwAV2. For the CV dataset, SII outperformed SwAV2 by 6 percentage points. The ARI and NMI values for the SwAV baseline models were low compared with those for the LLP-Co approaches for both scenarios, ranging from $-0.03$ to 0.50.  

All experiments reported relatively good performance with the kNN classifier. SI had the best results for both datasets. Note that the use of the global proportions from the census data resulted in a drop in performance compared to the use of the global proportions from the annotated dataset. This decrease in accuracy was expected because, first, the census data contains estimated proportions, not the exact ones, and second, we filtered the image by removing the regions with an NDVI standard deviation of less than 25\%, because the available optical images contain gaps due to the presence of clouds or acquisition problems. As a result, the agricultural mask may contain regions that are not plantations. As reported in \cite{9026754}, adjusting the percentage threshold for the standard deviation could lead to better results. A more reliable solution would train a classifier only to discriminate between agricultural and non-agricultural regions. However, this goes beyond the scope of this paper, and our objective is to show that without any fine-tuning and using only the estimated proportions, our methodology can achieve competitive performance. 

\begin{figure}[ht!]
    \centering
		\begin{subfigure}[b]{0.99\linewidth}
             \subfloat[\scriptsize{CV-SI}]{\includegraphics[width=0.25\linewidth]{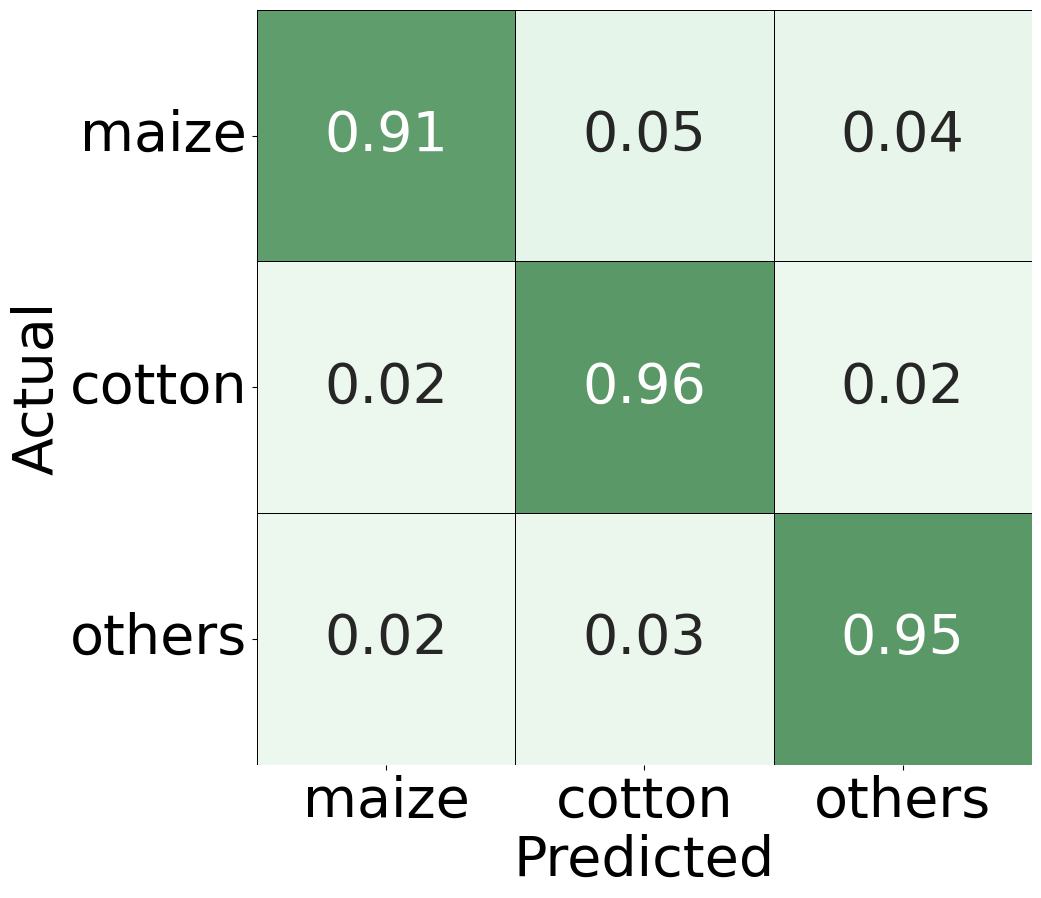}}
             \subfloat[\scriptsize{CV-SII}]{\includegraphics[width=0.25\linewidth]{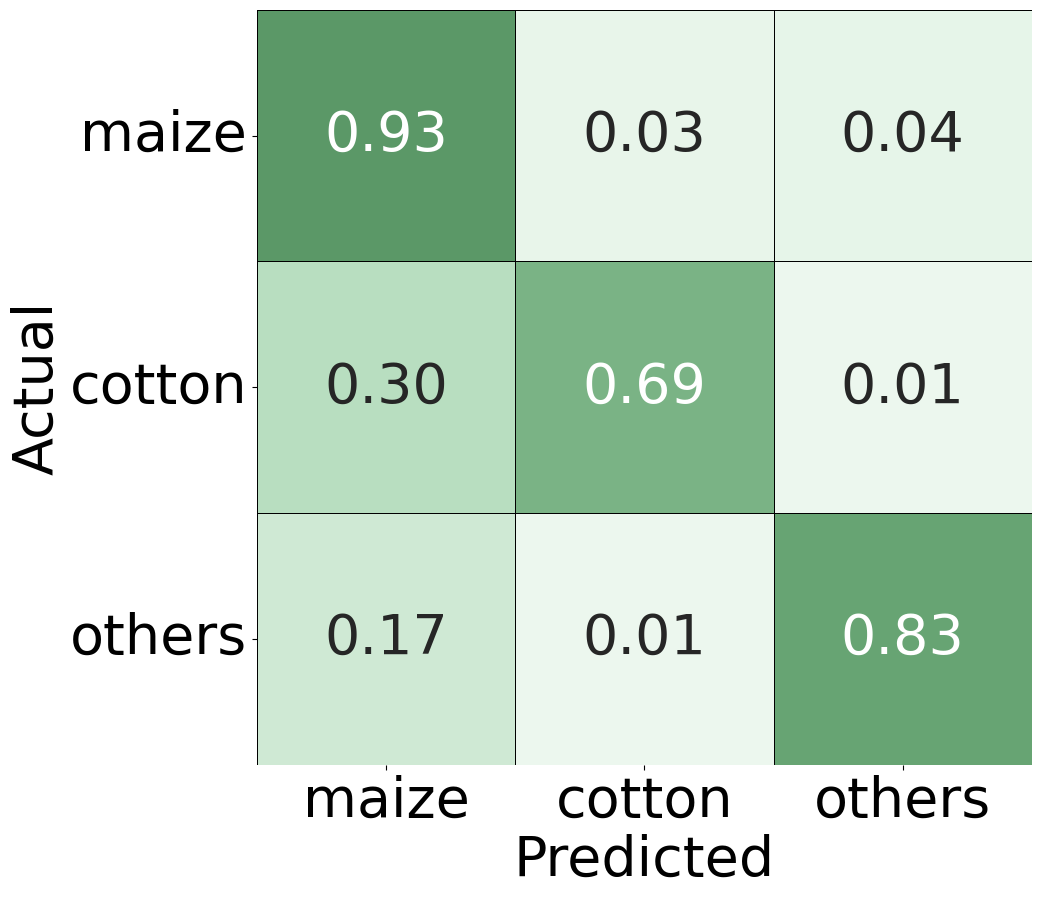}}
             \subfloat[\scriptsize{LEM-SI}]{\includegraphics[width=0.25\linewidth]{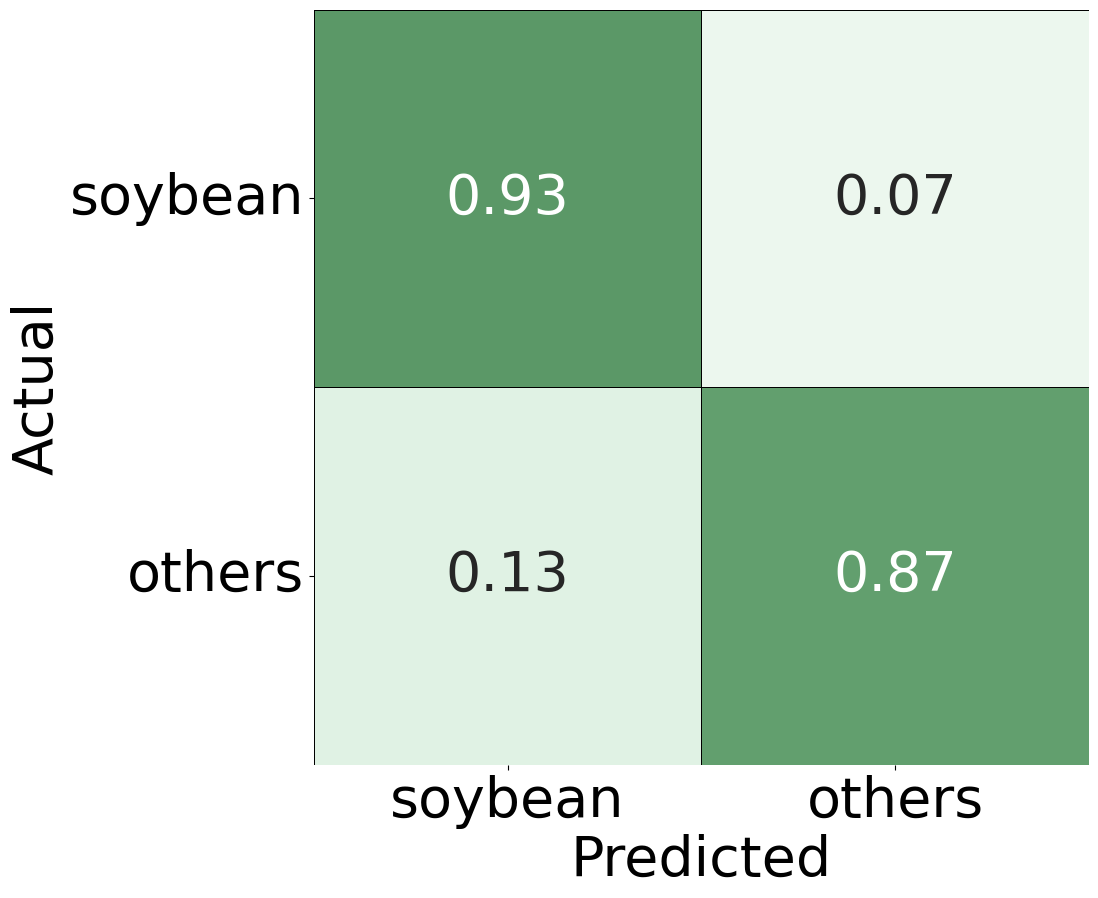}}
             \subfloat[\scriptsize{LEM-SII}]{\includegraphics[width=0.25\linewidth]{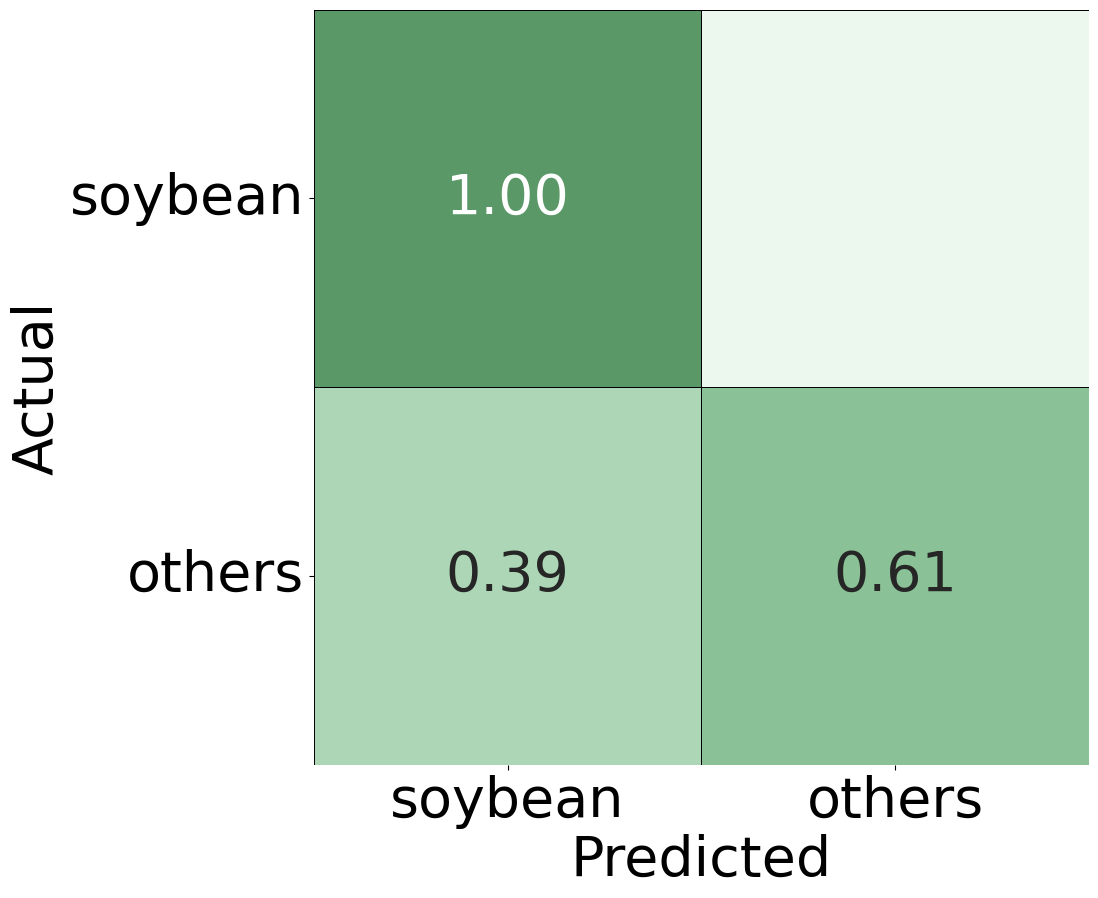}}
        \end{subfigure}
	\caption{Confusion matrices for scenarios SI and SII for the CV and LEM datasets evaluated for the test set and for major crops and class \textit{others}.}
	\label{fig:s1s2}
\end{figure}

Fig.~\ref{fig:s1s2} presents the confusion matrices for Table~\ref{tab:s1s2}. As expected, the per-class accuracy achieved higher performance for scenario SI, with values above 91\% for CV and above 87\% for LEM. For CV for scenario SII, 30\% of the \textit{cotton} samples were misclassified as \textit{maize}, and as discussed above, this result was not unexpected since the IBGE proportions and the dataset proportions are slightly different. For the LEM dataset, 40\% of class \textit{others} was misclassified as \textit{soybean} and the classification accuracy for \textit{soybean} was 100\%. Again, this result was not unexpected since the IBGE proportion for class \textit{soybean} surpasses the annotated dataset proportion by more than 17 percentage points.

\begin{figure*}[tb!]
\centering
\begin{adjustbox}{width=0.99\textwidth}
\begin{tabular}{r|c|c|c}
& Reference & SI & SII \\\toprule
{\rotatebox{90}{CV}}&\includegraphics[width=0.3\textwidth]{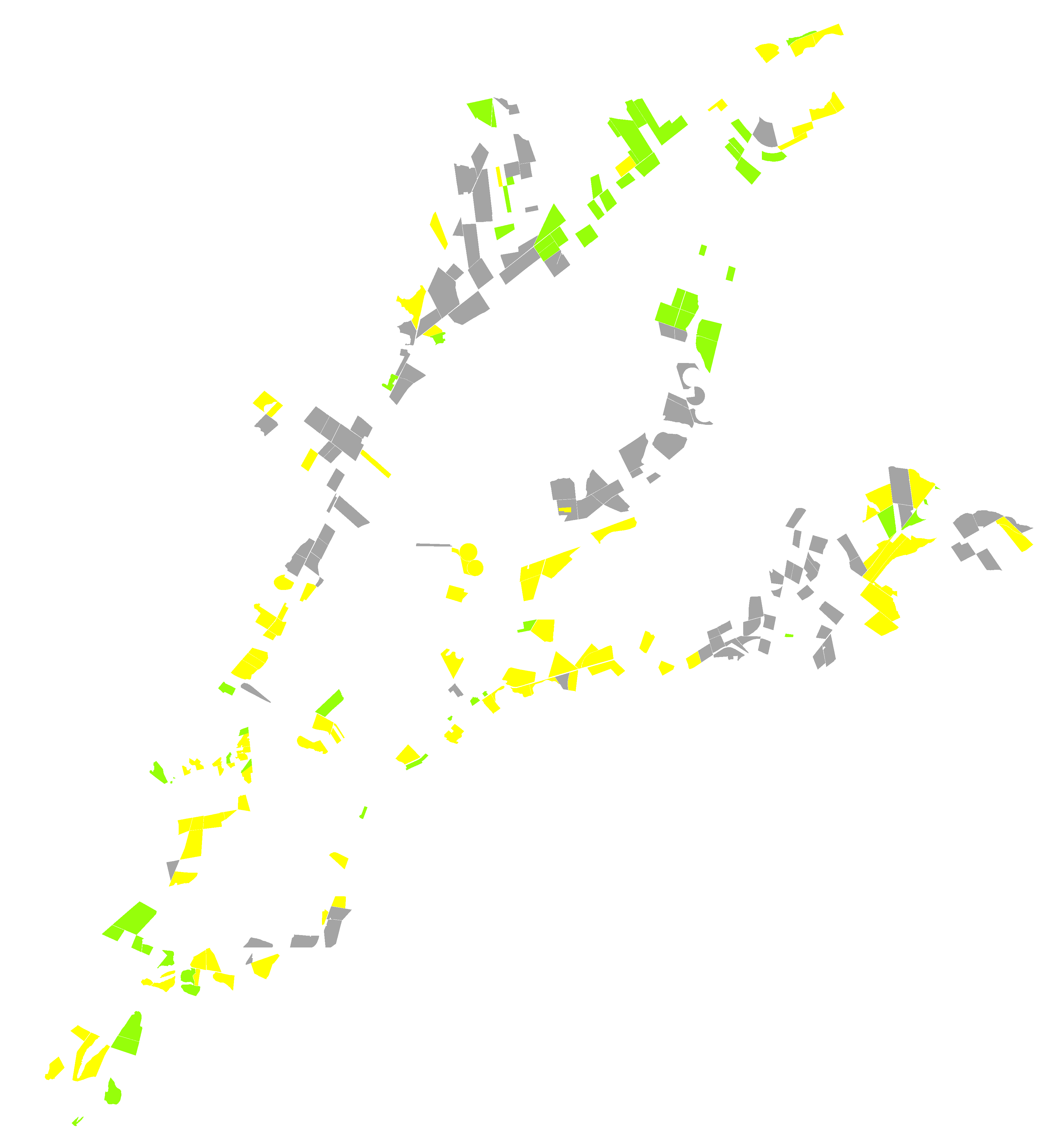}& \includegraphics[width=0.3\textwidth]{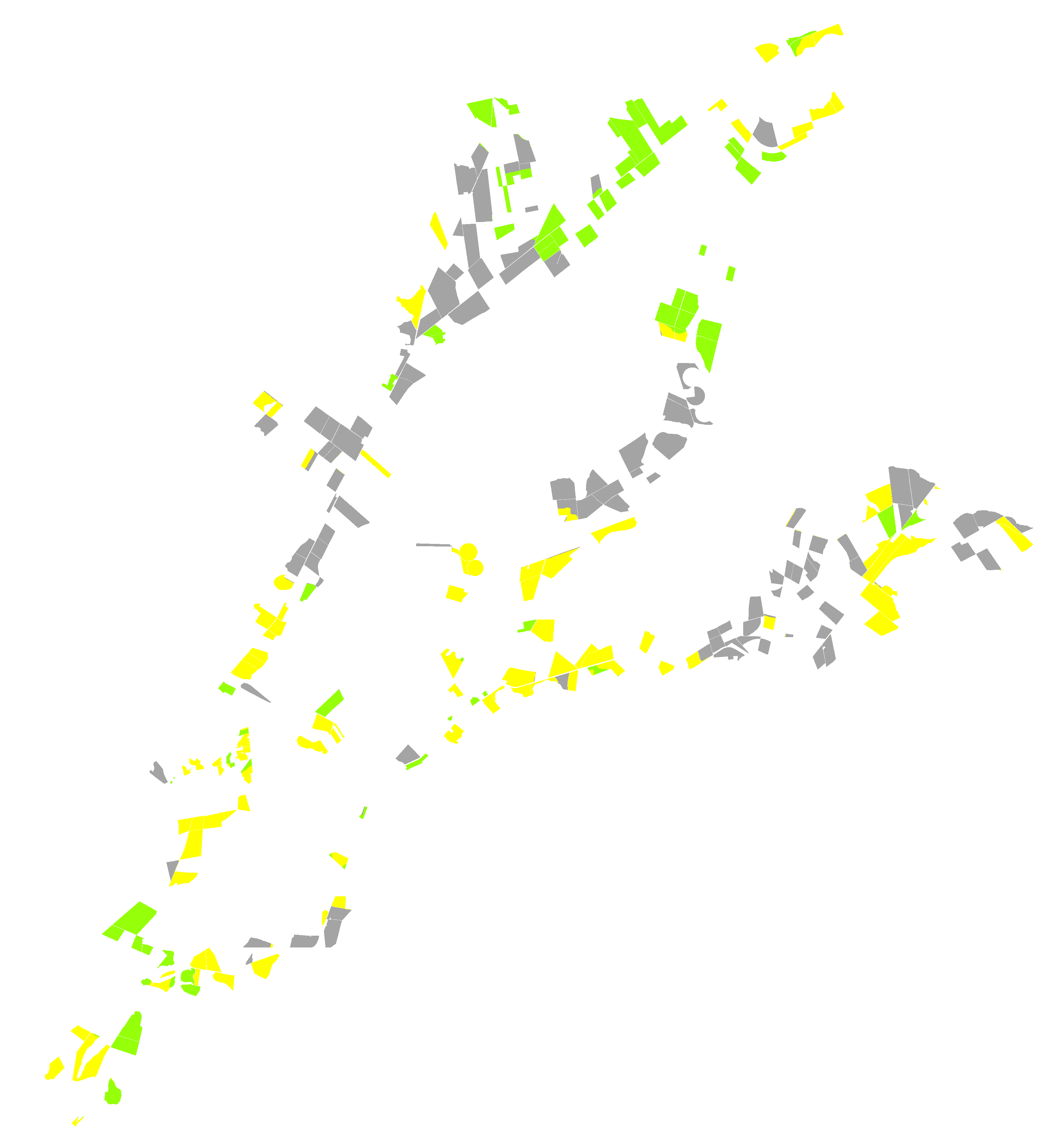}& \includegraphics[width=0.3\textwidth]{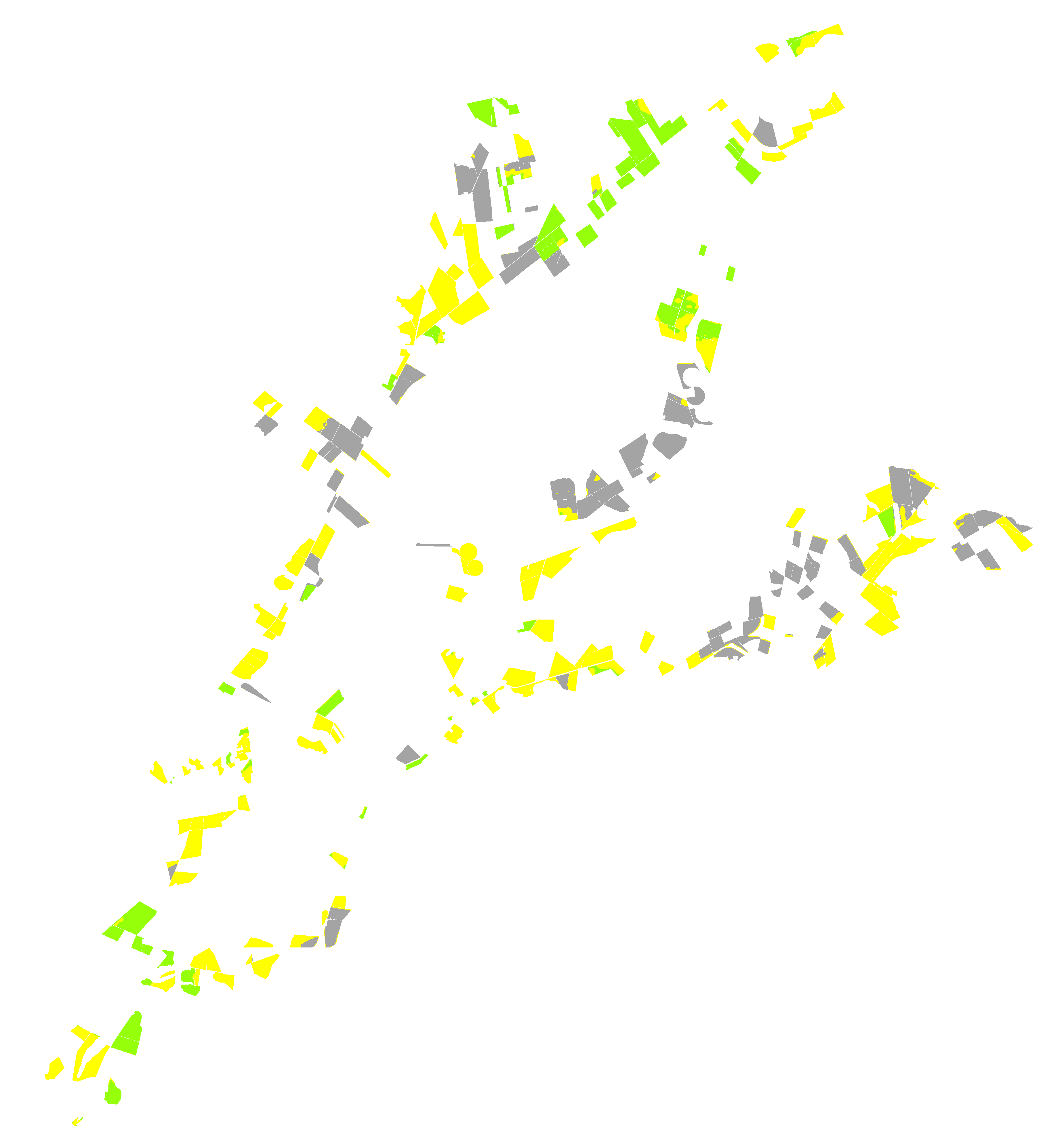}\\\midrule
{\rotatebox{90}{LEM}}&\includegraphics[width=0.3\textwidth]{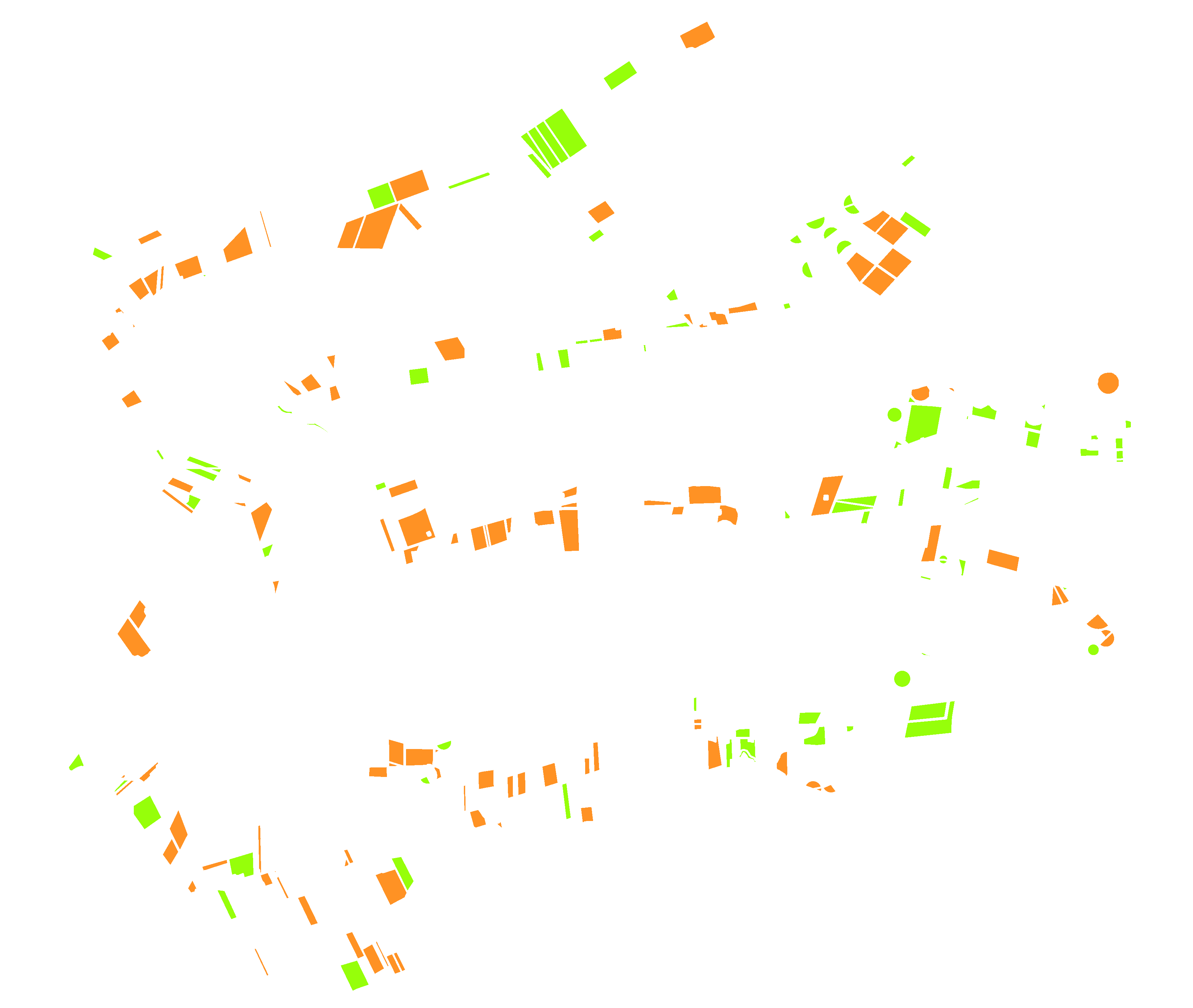}& \includegraphics[width=0.3\textwidth]{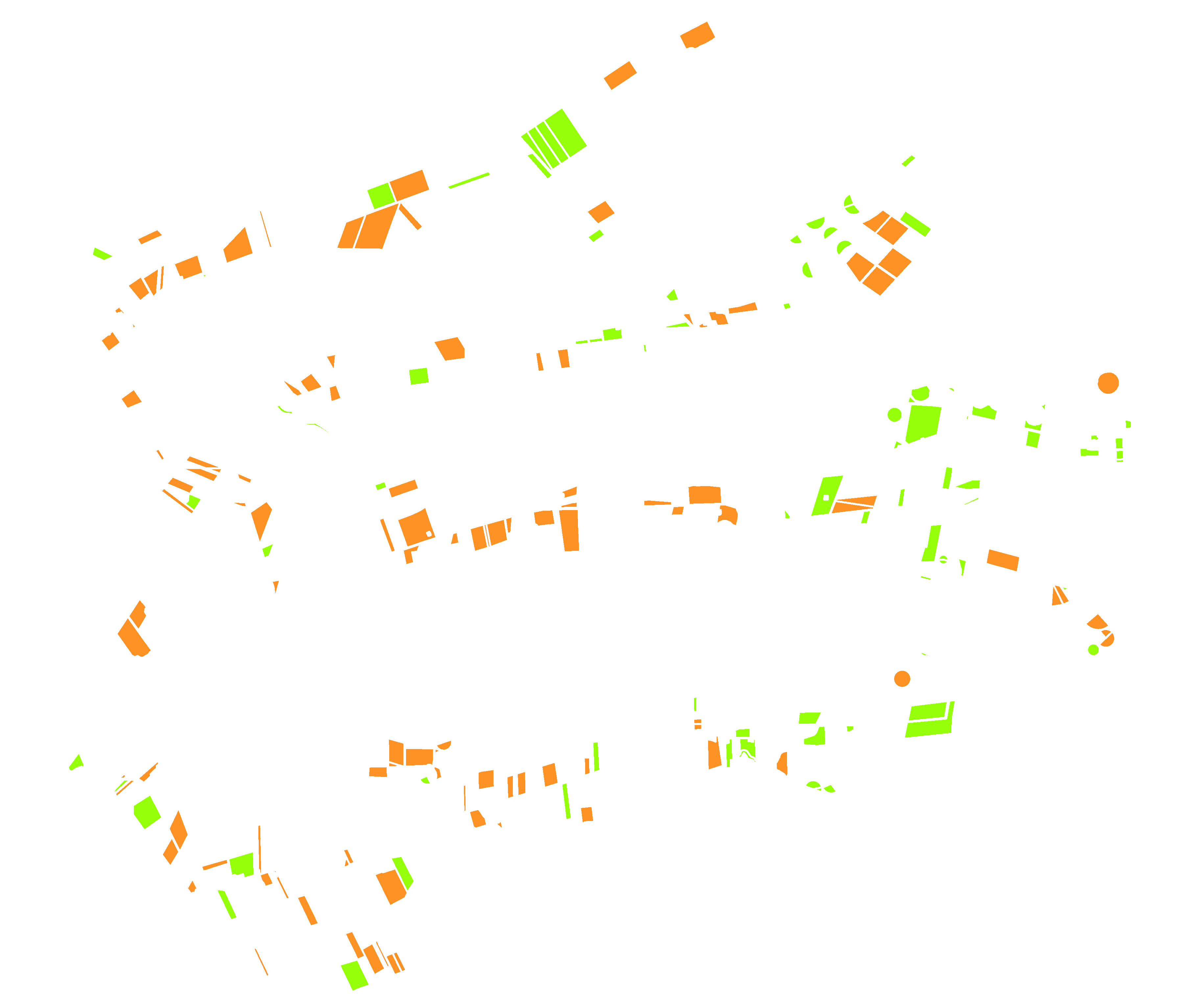}& \includegraphics[width=0.3\textwidth]{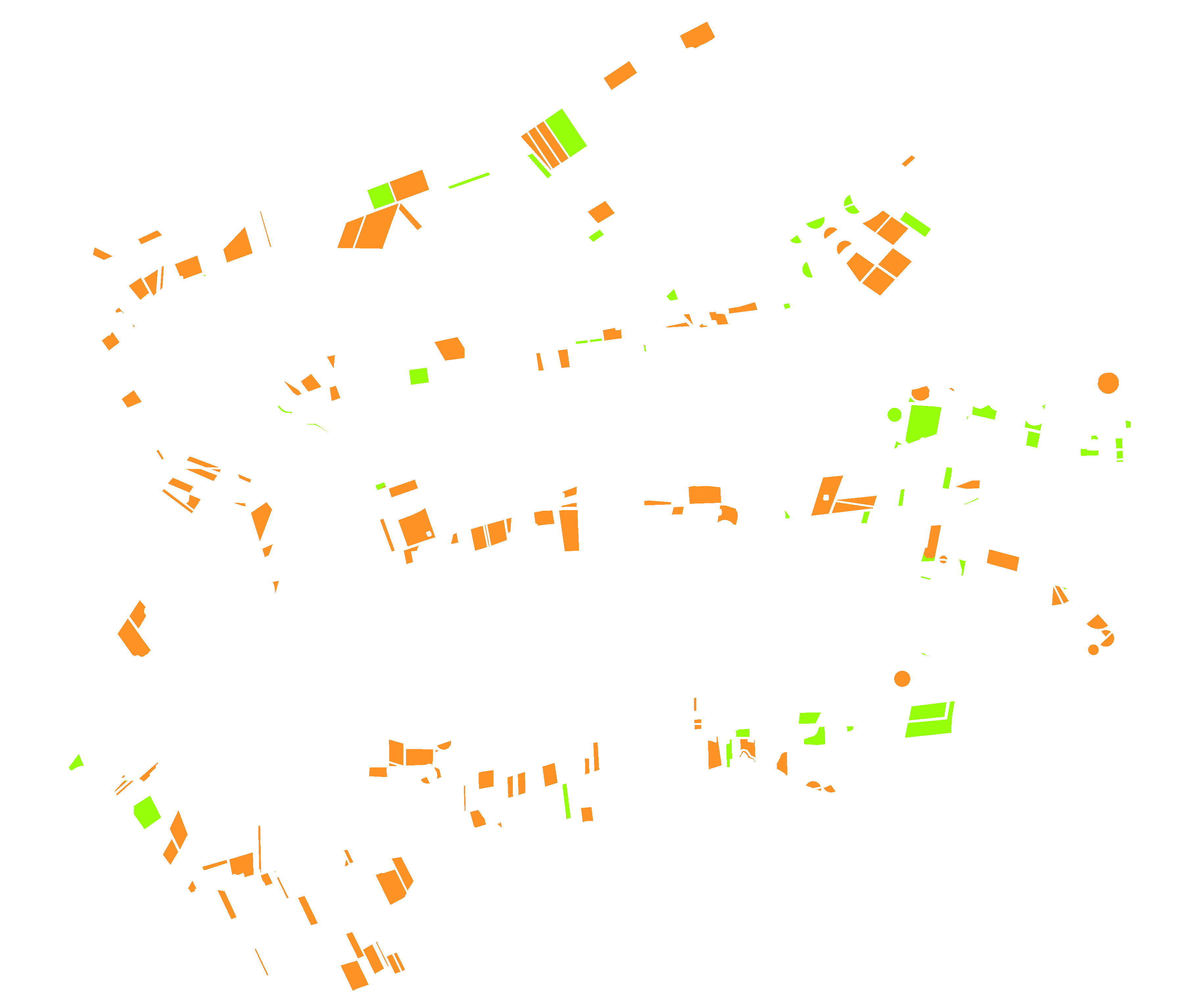}\\\bottomrule
\end{tabular}
\end{adjustbox}
\caption{Maps of the class output for SI and SII for the test region for the CV and LEM datasets. Crop types: \textit{soybean} (orange), \textit{maize} (yellow), \textit{cotton} (gray), and \textit{others} (green).}
\label{tab:clips1s2}
\end{figure*}

Fig.~\ref{tab:clips1s2} presents qualitative results for SI and SII. The classification for SI was better than that for SII. As observed for the confusion matrix, most SII errors are the misclassification of class \textit{cotton} as \textit{maize}. In addition, note the smoothness of the predictions for all methods, as most of the salt-and-pepper effect commonly observed in supervised approaches is not present. 

\subsection{Results for SIII and SIV}

Table~\ref{tab:s3s4} shows the performance with the test set for scenarios SIII and SIV and the baseline method. All classes in the datasets were used. In addition, we also used as a reference the fully supervised learning results reported for the CV dataset on the same date using optical data \cite{9554011} and for the LEM dataset on the same date using SAR data \cite{martinez2021fully}. These supervised results were obtained using a fully convolutional network with the ResNet backbone for the CV dataset and 3D bidirectional ConvLSTM for the LEM dataset. For more details of these architectures, see \cite{9554011} and \cite{martinez2021fully}. 

\begin{table*}[ht]
    \centering
        \caption{Test performance for CV for scenarios SIII and SIV and the baseline SwAV3 method with different bag sizes.}        
        \label{tab:s3s4}
        \begin{tabular}{c c cccccc cccc cccc c}
        \toprule
        \multirow{2}{*}{Dataset} & \multirow{2}{*}{Metric} & \multicolumn{6}{c}{SIII} & \multicolumn{4}{c}{SIV} & \multicolumn{4}{c}{SwAV3} & \multirow{2}{*}{Sup} \\
        \cmidrule(lr){3-8}
        \cmidrule(lr){9-12}
        \cmidrule(lr){13-16}
        & & 32 & 64 & 128 & 256 & 512 & 1024 & 256 & 512 & 1024 & 2048 & 256 & 512 & 1024 & 2048 & \\\midrule
        \multirow{4}{*}{CV} & $Acc_P$ & 82.3 & 85.7 & 86.4 & 83.2 & 85.1 & 84.4 & 73.8 & 76.4 & 75.5 & 75.1 & -- & -- & -- & -- & \multirow{3}{*}{89.1}\\
        & $Acc_H$ & 82.3 & 85.7 & 86.4 & 83.2 & 85.1 & 84.4 & 77.9 & 80.1 & 80.2 & 79.5 & 46.4 & 37.7 & 40.0 & 42.1\\
        & kNN & 86.7 & 87.3 & 85.7 & 86.0 & 85.5 & 86.8 & 86.4 & 88.6 & 87.5 & 88.4 & 88.4 & 85.7 & 86.4 & 86.5 &\\
        & ARI & 0.64 & 0.70 & 0.71 & 0.67 & 0.71 & 0.74 & 0.65 & 0.75 & 0.78 & 0.78 & 0.19 & 0.10 & 0.19 & 0.24 & -- \\
        & NMI & 0.57 & 0.60 & 0.63 & 0.56 & 0.61 & 0.62 & 0.60 & 0.63 & 0.65 & 0.64 & 0.33 & 0.34 & 0.39 & 0.46 & -- \\\midrule
        \multirow{4}{*}{LEM} & $Acc_P$ & 95.3 & 94.2 & 86.9 & 92.7 & 57.8 & 62.1 & 52.6 & 51.1 & 53.8 & 58.0 & -- & -- & -- & -- & \multirow{3}{*}{93.1}\\
        & $Acc_H$ & 95.3 & 94.2 & 87.2 & 92.7 & 69.9 & 79.1 & 62.8 & 61.2 & 66.2 & 68.8 & 40.9 & 36.5 & 34.5 & 35.3\\
        & kNN & 95.1 & 94.6 & 95.0 & 94.9 & 93.1 & 90.3 & 89.7 & 90.5 & 91.9 & 92.0 & 91.2 & 91.7 & 91.7 & 92.9 \\
        & ARI & 0.94 & 0.91 & 0.85 & 0.87 & 0.53 & 0.73 & 0.35 & 0.34 & 0.51 & 0.60 & 0.13 & 0.09 & 0.12 & 0.13\\
        & NMI & 0.87 & 0.84 & 0.74 & 0.80 & 0.57 & 0.61 & 0.49 & 0.53 & 0.55 & 0.55 & 0.45 & 0.38 & 0.41 & 0.40\\\bottomrule
    \end{tabular}
\end{table*}

As expected, SIII delivers the best results, with accuracy values ranging from 82\% to 86\% for CV, close to the overall accuracy of 89\% for the fully supervised approach. Like CV, SIII for LEM had impressive results for bag size from 32 to 256, with accuracies from 86.9\% to 95\% and outperforming the supervised approach for bag sizes 32 and 64. For this scenario for CV, observe that $Acc_P$ and $Acc_H$ had the same values, which indicates that, when working with optical data, using the exact bag proportion means samples are mapped to the correct cluster without the typical cluster swapping effect. However, for the LEM dataset, $Acc_P < Acc_H$ from some of the bags. Unlike CV, we observed more variation in the classification performance for the LEM dataset as the bag size increased. These results indicate that the model seems less robust with SAR images than with optical images. Nonetheless, the decrease in performance for larger bag sizes (512 and 1024 in Table \ref{tab:s3s4}) is a common problem that has been reported in previous research on LLP \cite{liu2019learning}. We observe similar variation for ARI and NMI for both datasets, indicating that the model is sensitive to unbalanced datasets. 

\begin{figure}[ht!]
    \centering
		\begin{subfigure}[b]{0.99\linewidth}
             \centering
             \subfloat[\scriptsize{CV-SIII-128}]{\includegraphics[width=0.49\linewidth]{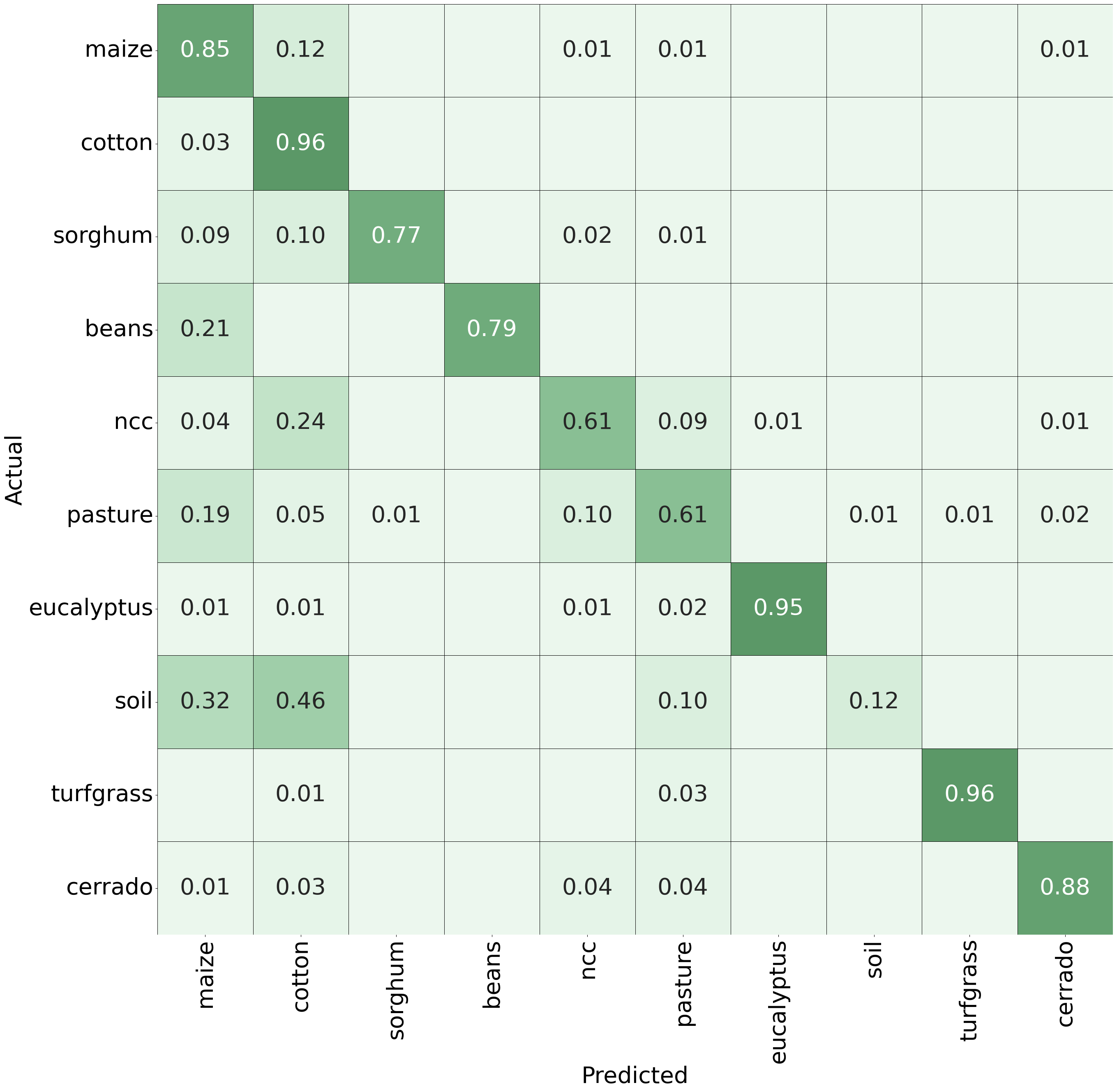}}
             \subfloat[\scriptsize{CV-SIII-256}]{\includegraphics[width=0.49\linewidth]{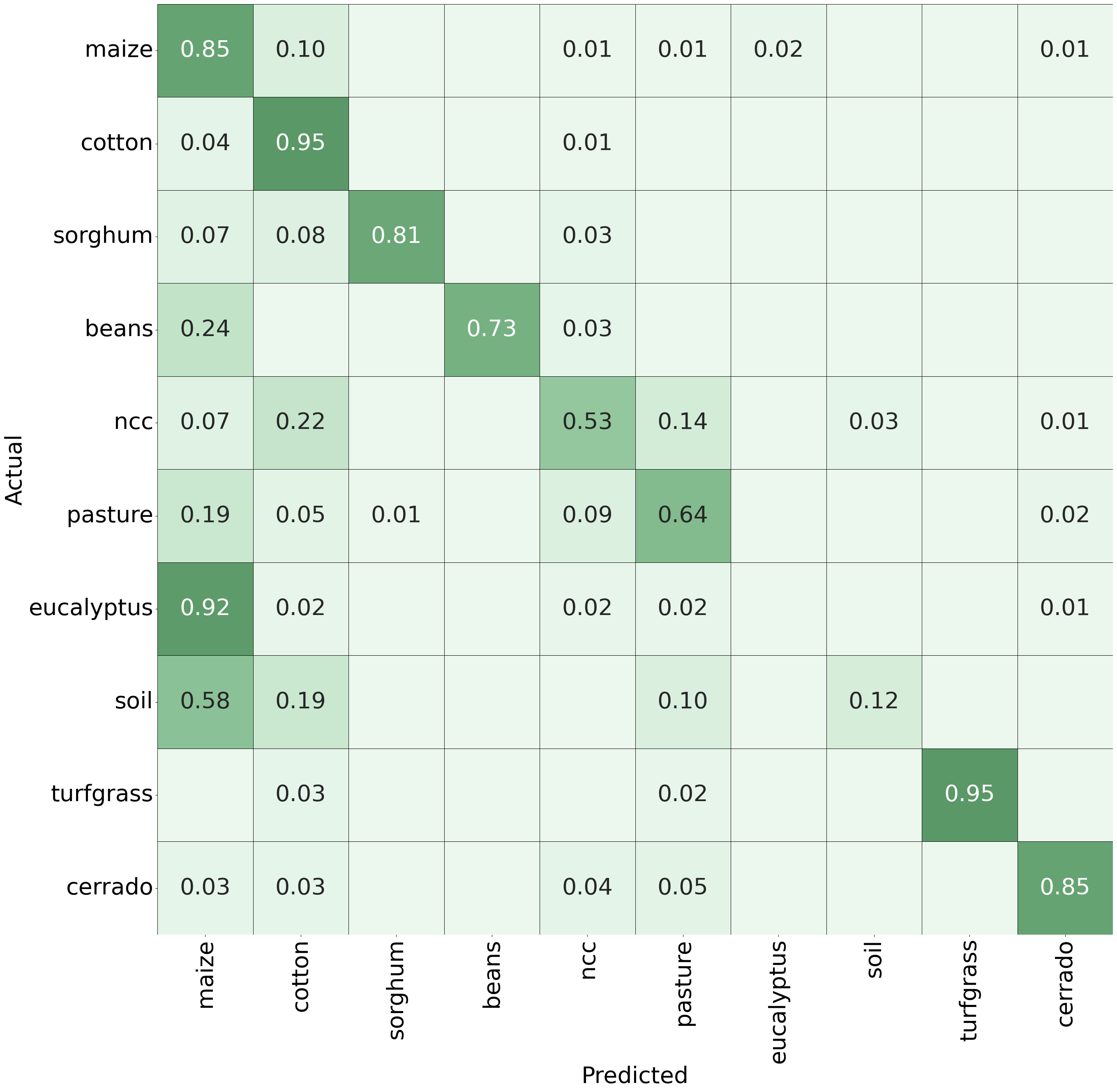}}\\
             \subfloat[\scriptsize{LEM-SIII-128}]{\includegraphics[width=0.49\linewidth]{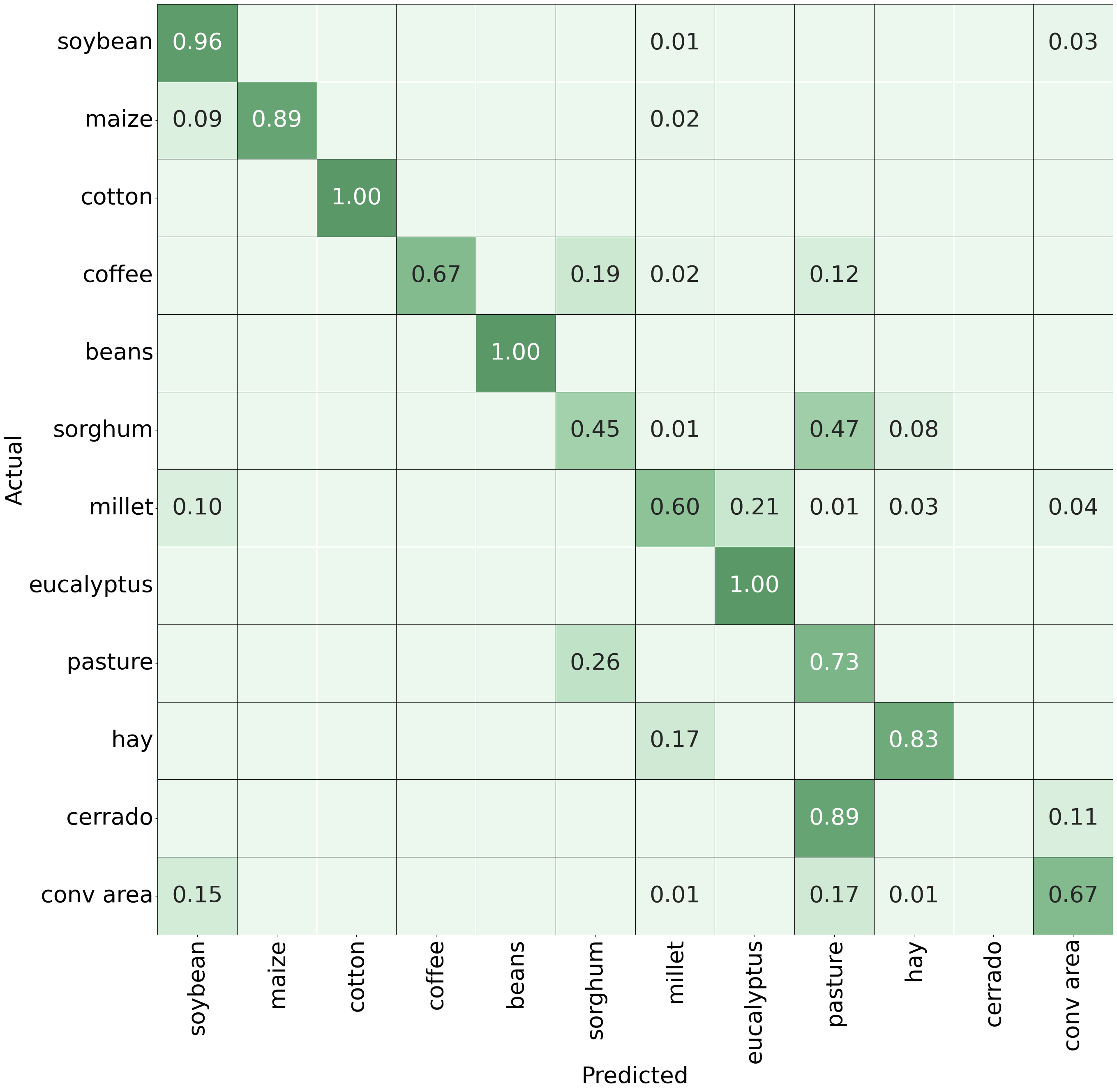}}
             \subfloat[\scriptsize{LEM-SIII-256}]{\includegraphics[width=0.49\linewidth]{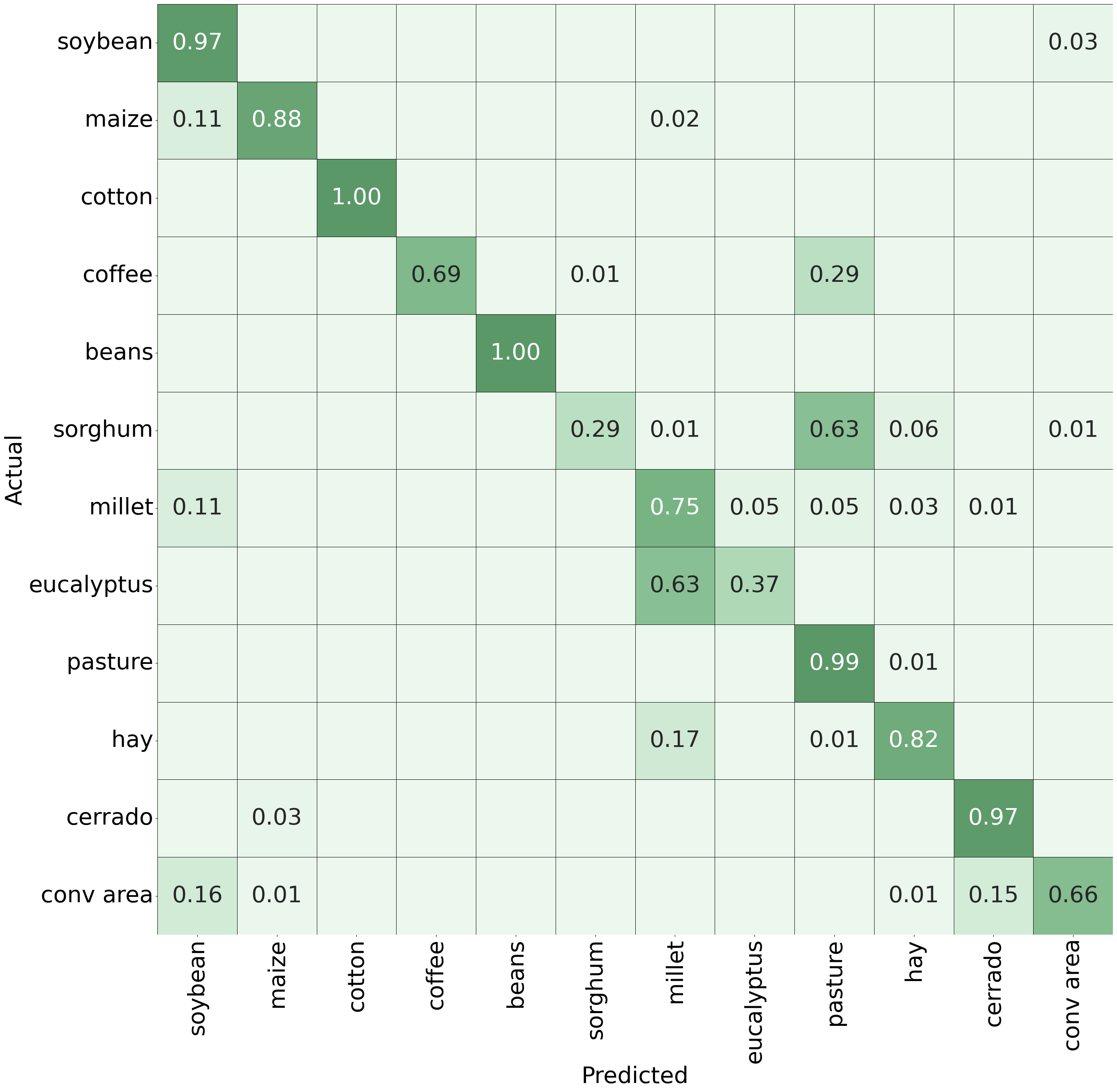}}       
        \end{subfigure}
	\caption{Confusion matrices for scenario SIII for bag sizes of 128 and 256 evaluated for the test set for the CV and LEM datasets.}
	\label{fig:confs3}
\end{figure}

To better understand this behavior, Fig.~\ref{fig:confs3} shows the confusion matrices for bag sizes of 128 and 256 for the CV and LEM datasets. Note that the accuracies for the two datasets for each of the major crop types (\textit{soybean}, \textit{maize}, and \textit{cotton}) are roughly the same for each bag size. For the other classes, the best performance was either for bag 128 or for bag 256, except for class \textit{eucalyptus} for bag size 256 in CV and for class \textit{cerrado} for bag size 128 in LEM, which were entirely misclassified.

For scenario SIV, there was an increase in performance with an increase of the bag size. In terms of accuracy, ARI, and NMI, the best performance was for bag size 1024 for CV and for bag size 2048 for LEM. Nonetheless, for CV, the performance for bag size 2048 was very similar to that for bag size 1024. The swap cluster phenomenon ($Acc_P < Acc_H$) also occurs for this scenario, which was somewhat expected since the proportions for some of the minor classes are very similar with very low values, for example \textit{eucalyptus}/\textit{cerrado}, \textit{sorghum}/\textit{soil}, and \textit{beans}/\textit{turfgrass} for CV  and \textit{eucalyptus}/\textit{hay}/\textit{coffee}, \textit{soil}/\textit{not identified}, and \textit{cotton}/\textit{pasture} for LEM (Fig.~\ref{fig:boxplot}). 

\begin{figure}[ht!]
    \centering
		\begin{subfigure}[b]{0.99\linewidth}
             \centering
             \subfloat[\scriptsize{CV-SIV-1024}]{\includegraphics[width=0.49\linewidth]{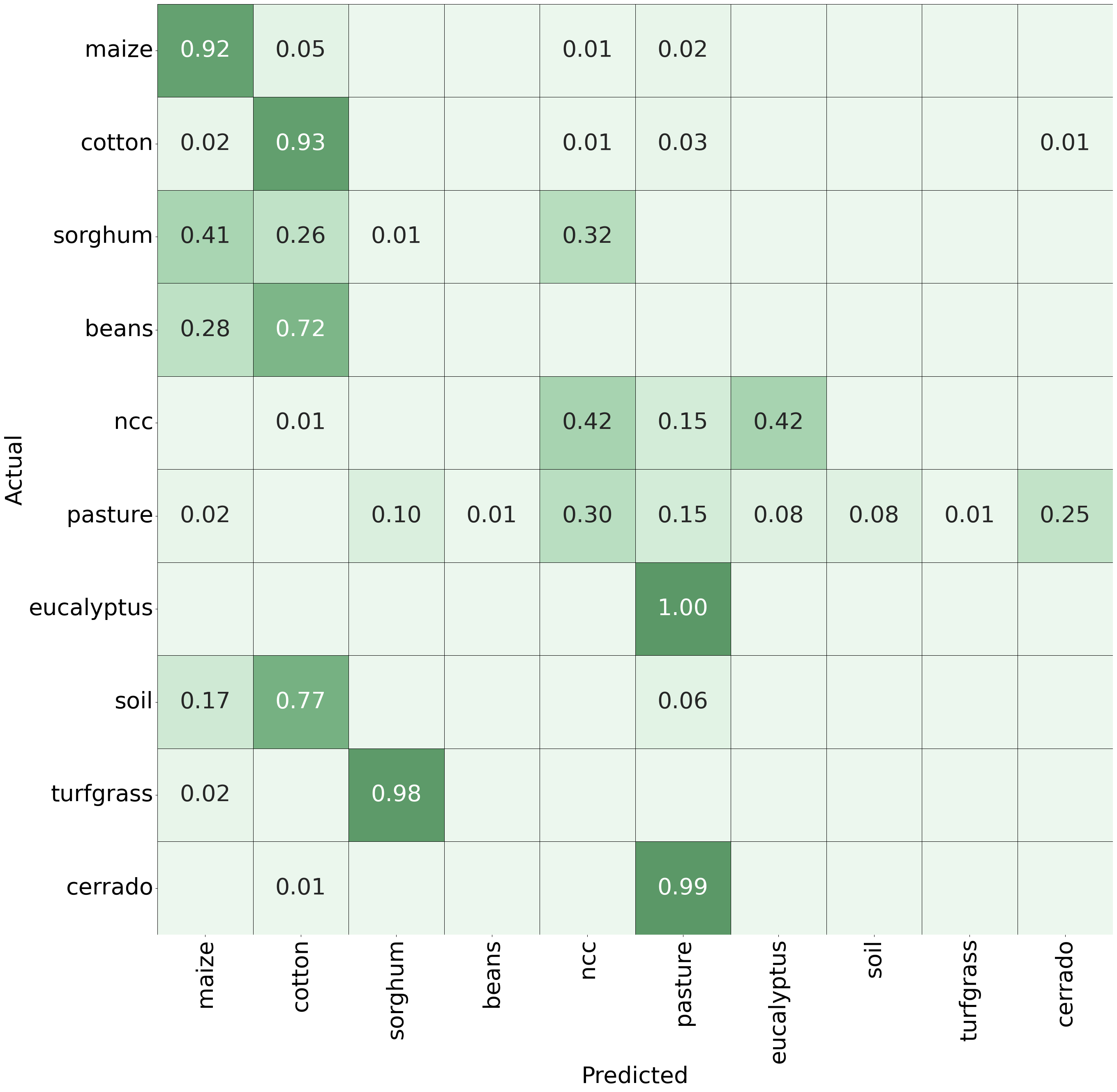}}
             \subfloat[\scriptsize{LEM-SIV-2048}]{\includegraphics[width=0.49\linewidth]{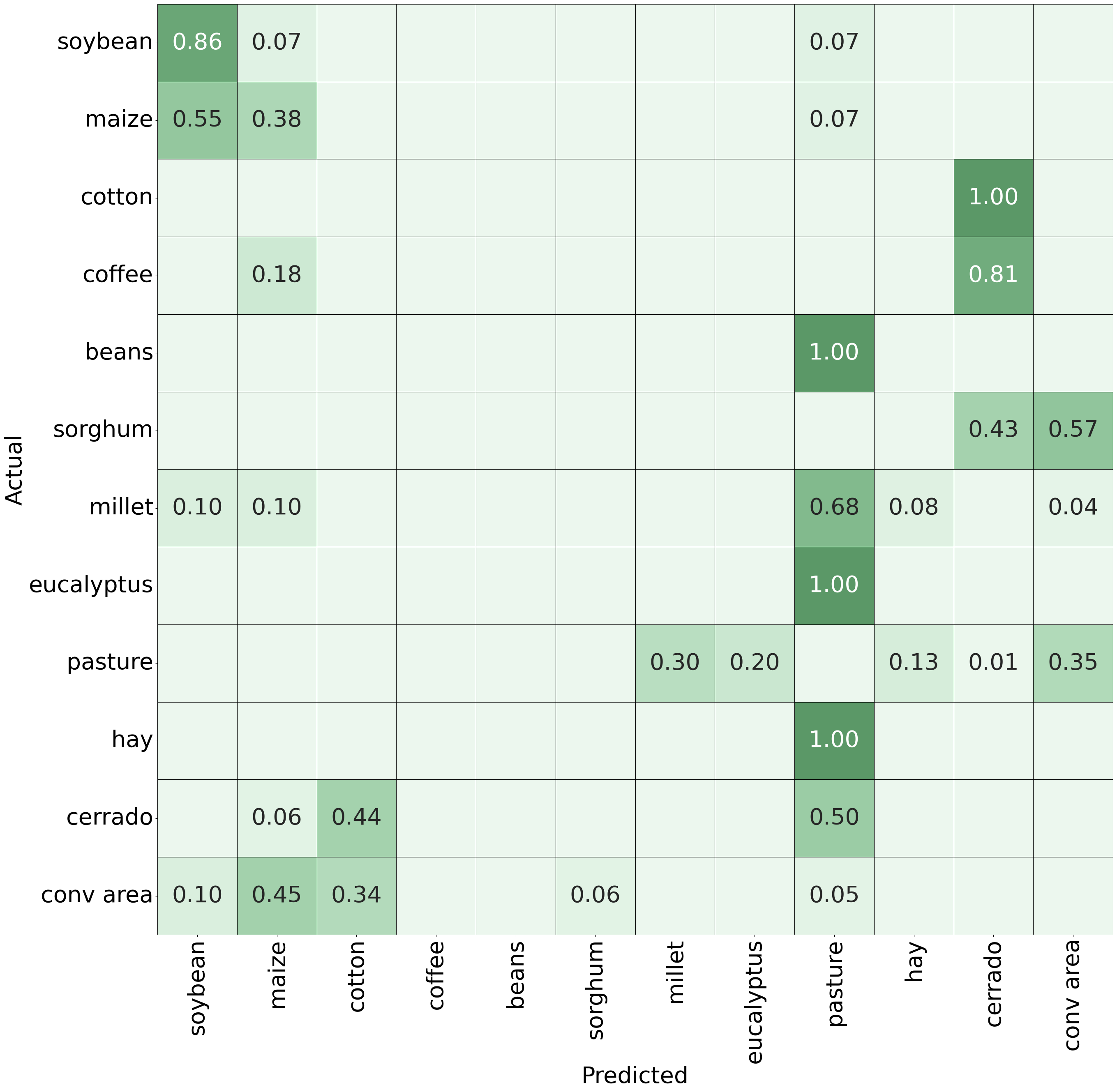}}
        \end{subfigure}
	\caption{Confusion matrices for scenario SIV for bag sizes 1024 and 2048 evaluated for the test set for the CV and LEM datasets, respectively.}
	\label{fig:cvconfs2}
\end{figure}

Fig.~\ref{fig:cvconfs2} shows the confusion matrices for bag sizes of 1024 and 2048 that reported the best results for the CV and LEM datasets, respectively. The clusters were assigned with the Hungarian algorithm. For both datasets, most of the minor classes were completely misclassified, like \textit{sorghum}, \textit{beans}, \textit{soil}, \textit{soil}, \textit{coffee}, and \textit{hay}. Still, note that the major crop types for both datasets were correctly classified without the cluster swap problem. 

\begin{figure}[ht!]
    \centering
		\begin{subfigure}[b]{0.99\linewidth}
             \subfloat[\scriptsize{CV-256}]{\includegraphics[width=0.24\linewidth]{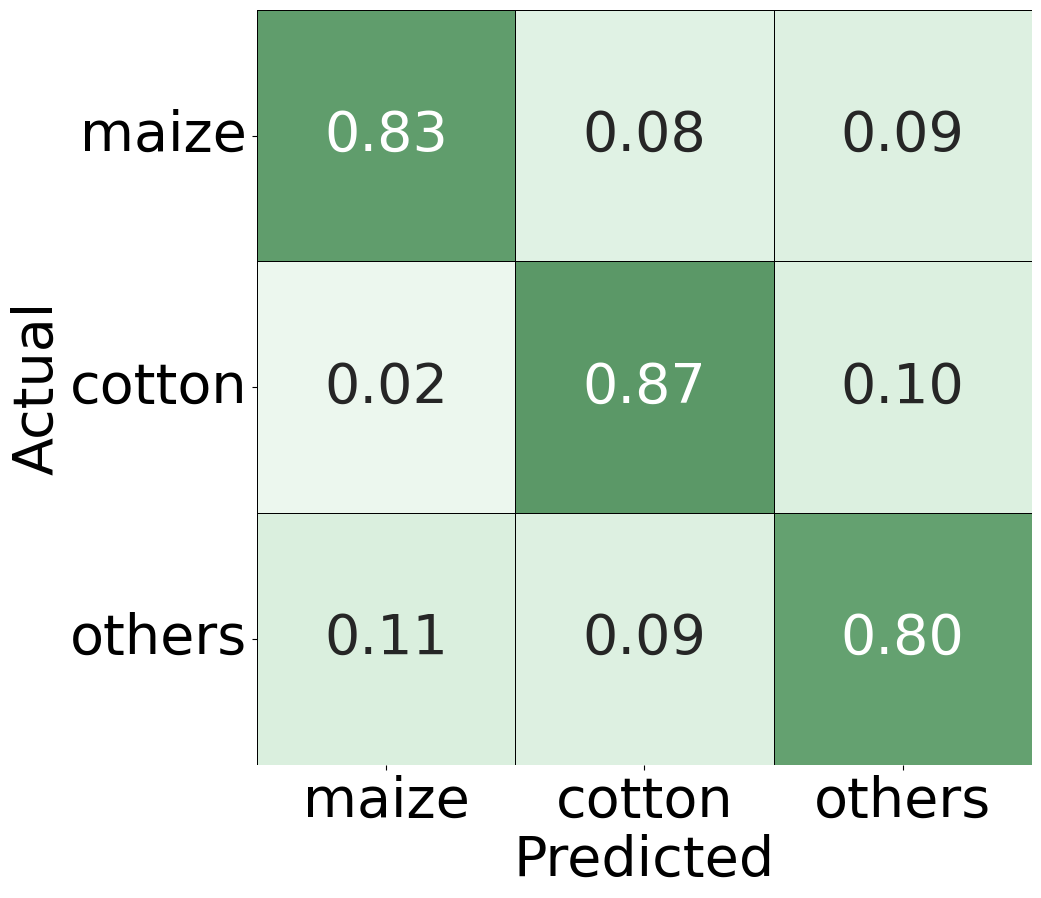}}
             \subfloat[\scriptsize{CV-512}]{\includegraphics[width=0.24\linewidth]{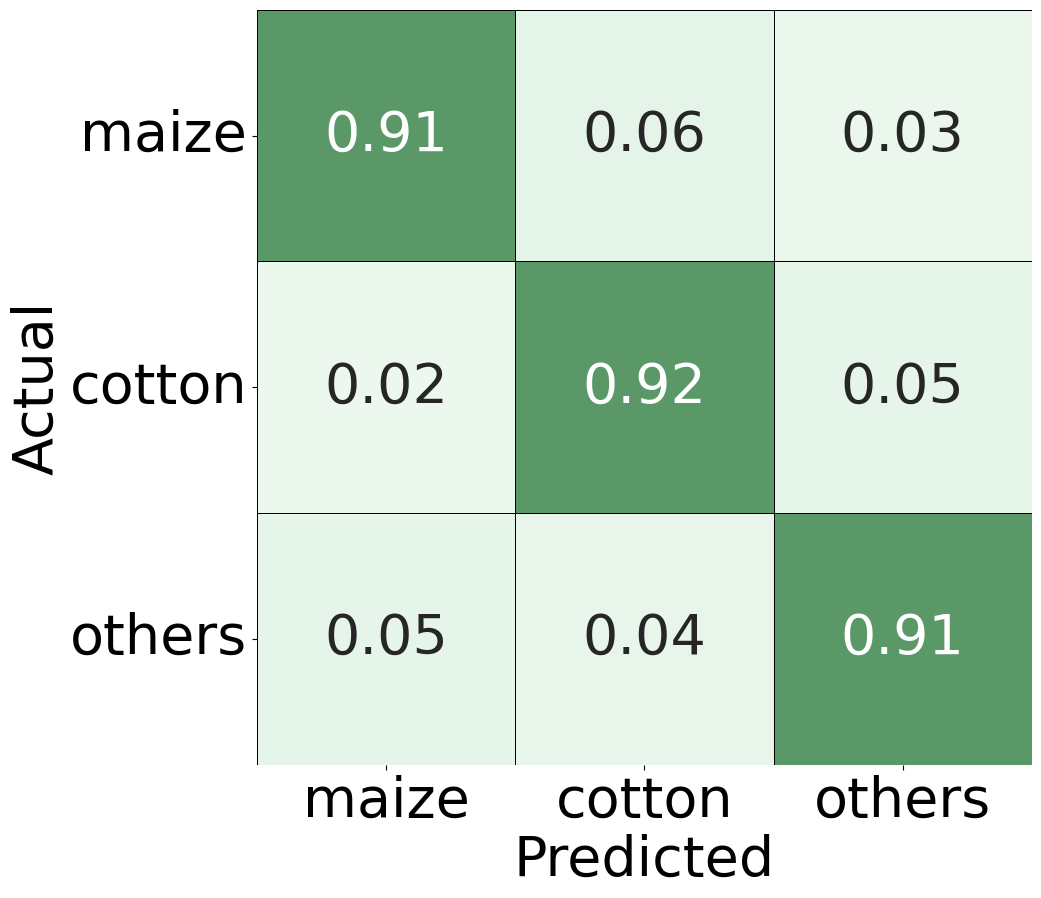}}
             \subfloat[\scriptsize{CV-1024}]{\includegraphics[width=0.24\linewidth]{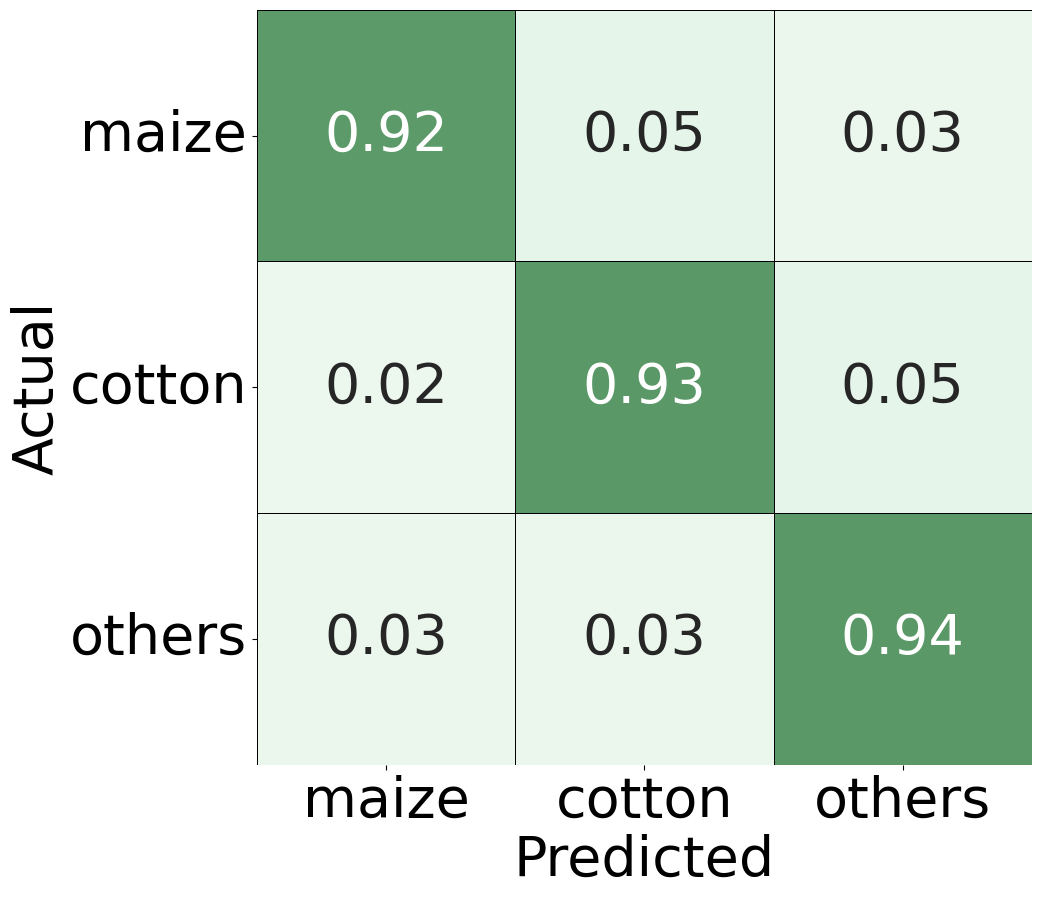}}
             \subfloat[\scriptsize{CV-2048}]{\includegraphics[width=0.24\linewidth]{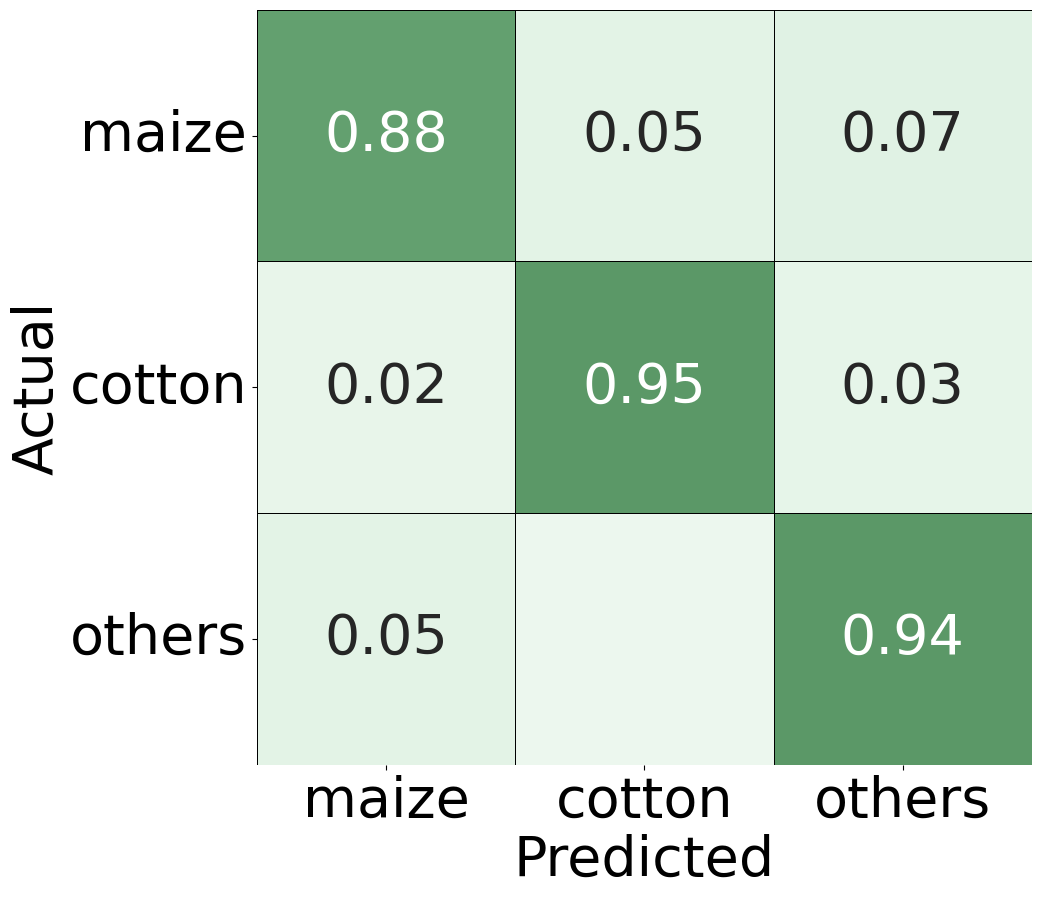}}\\
             \subfloat[\scriptsize{LEM-256}]{\includegraphics[width=0.24\linewidth]{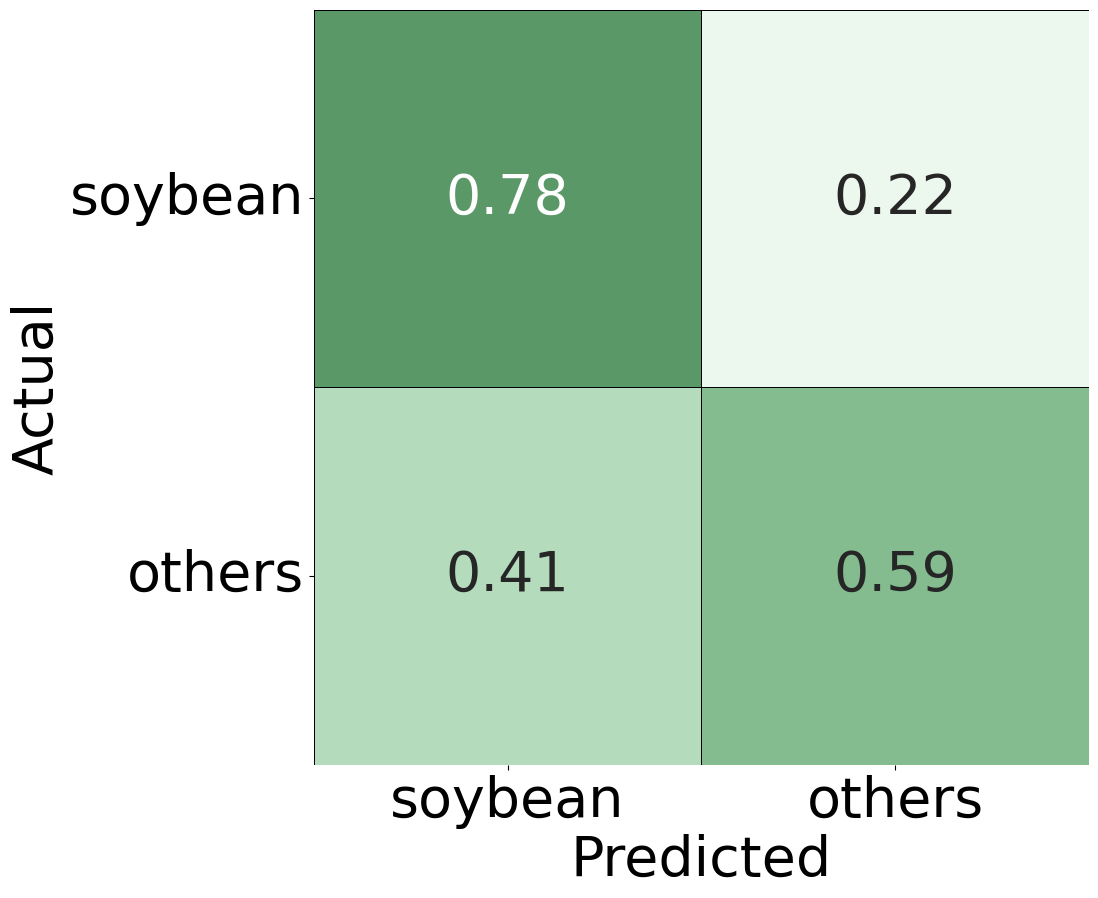}}
             \subfloat[\scriptsize{LEM-512}]{\includegraphics[width=0.24\linewidth]{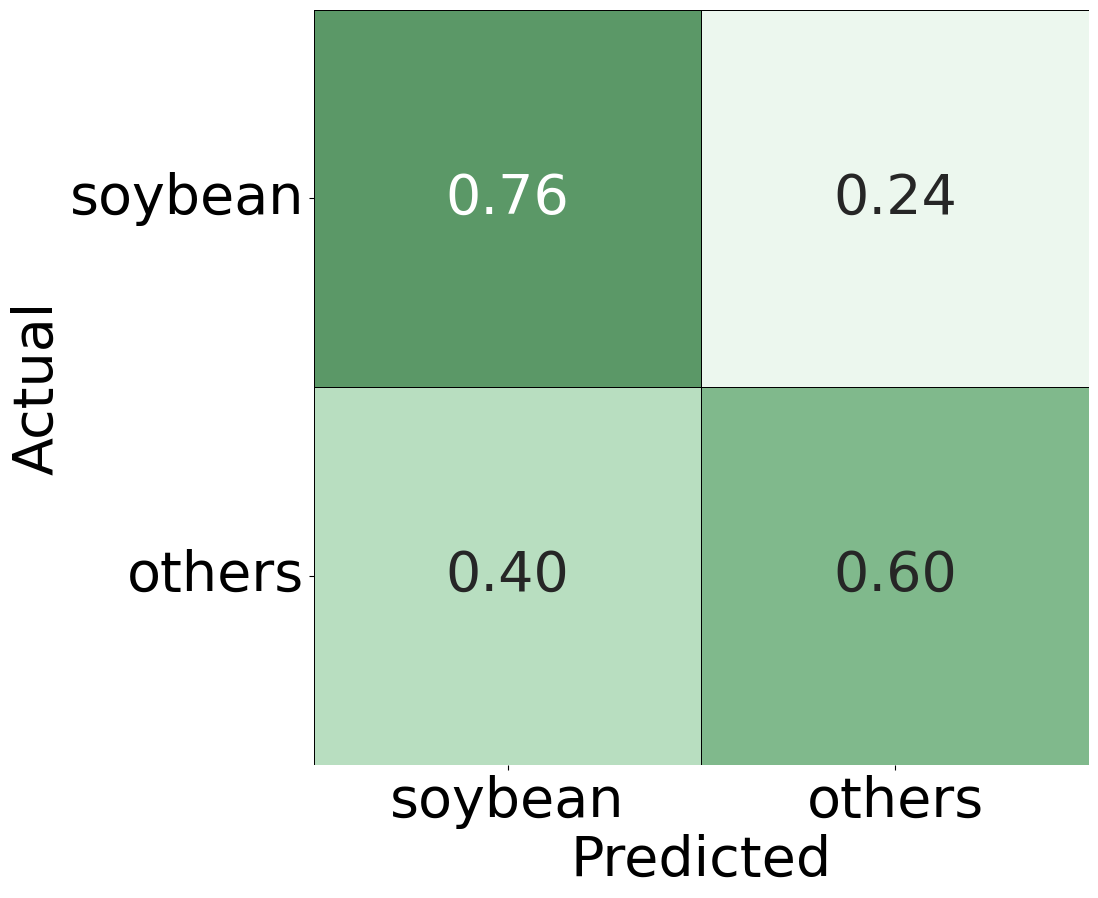}}
             \subfloat[\scriptsize{LEM-1024}]{\includegraphics[width=0.24\linewidth]{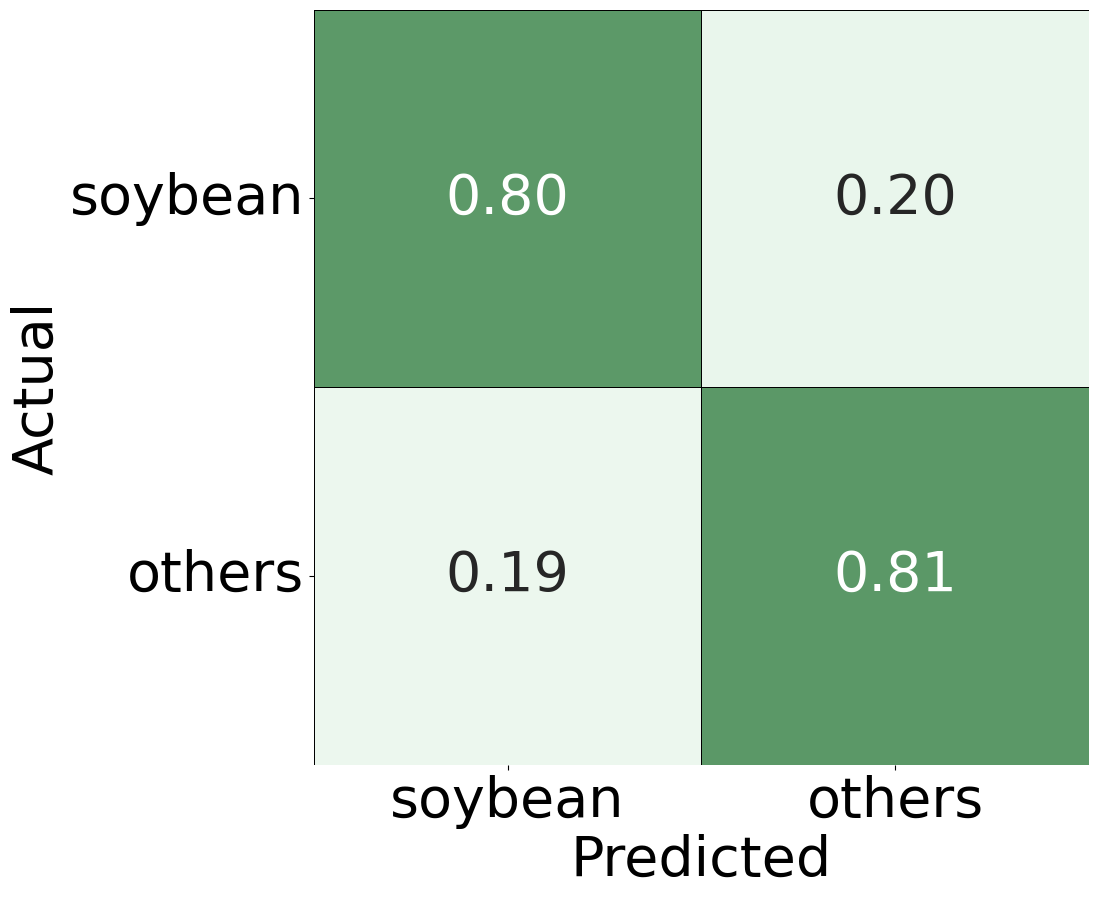}}
             \subfloat[\scriptsize{LEM-2048}]{\includegraphics[width=0.24\linewidth]{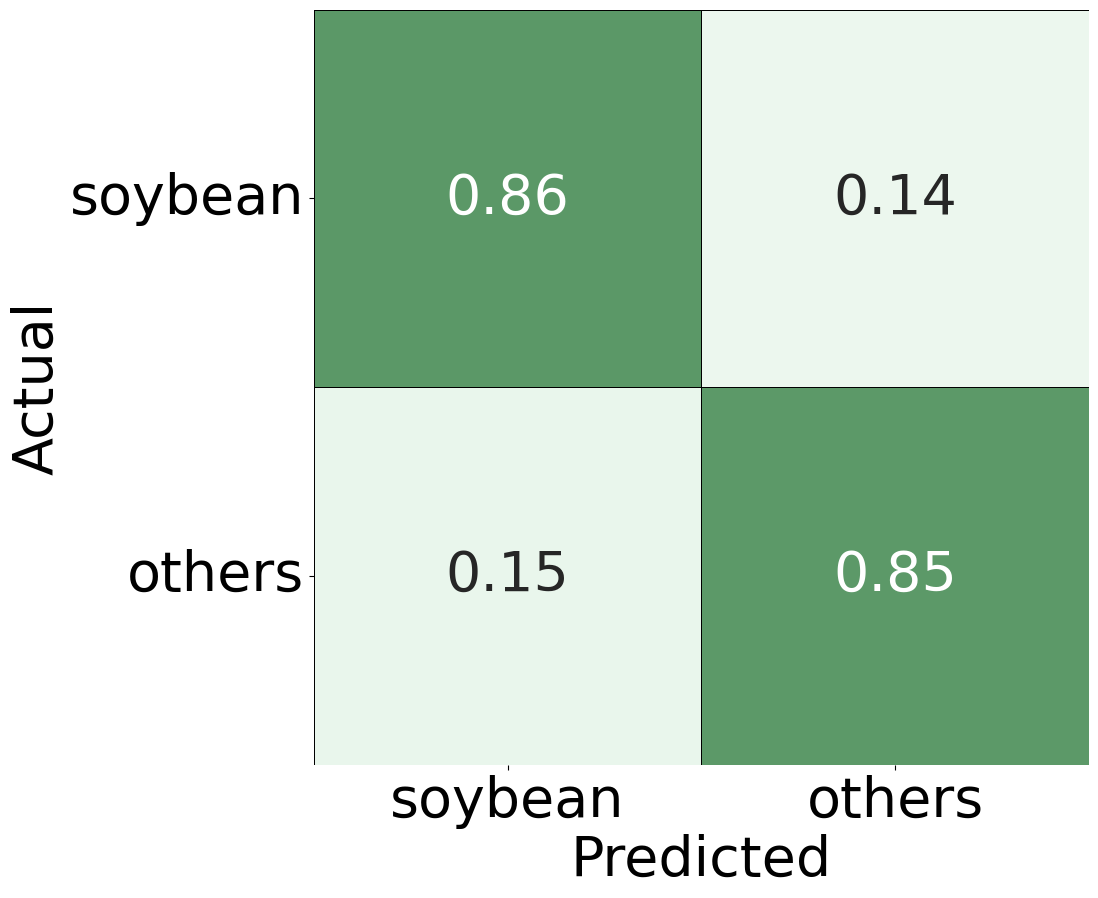}}
        \end{subfigure}
	\caption{Confusion matrices for scenario SIV for bag sizes of 256, 512, 1024, and 2048 evaluated for the test set for major crops and class \textit{others} for both datasets.}
	\label{fig:confs4}
\end{figure}

These results indicate that there was a strong dependency on the vector of proportions, most likely related to the highly unbalanced nature of the datasets. Nonetheless, for the major crops, the model achieved good accuracy for all bag sizes. Fig.~\ref{fig:confs4} presents the confusion matrices for each bag size for the major crop types for each dataset. The other crop types are grouped in class \textit{others}. For the LEM dataset, the precision of the classification increased proportionally with the bag size, achieving improvements of up to 8 percentage points for class \textit{soybean} and 26 percentage points for class \textit{others} when comparing bag sizes 256 and 2048. Smaller gains were observed for the CV dataset, with the most significant improvement when changing from bag size 256 to 512. Since we used SAR data for the LEM dataset and since LEM has more classes than CV, the LEM model was expected to need larger bag sizes to converge.

Comparing SIII and SIV with the baseline SwAV3 method highlights the limitations of training a fully unsupervised approach with such complex datasets with a large number of classes. The SwAV3 method reported deficient performance for all bag sizes for this experimental setup.

\begin{figure*}[ht!]
\centering
\begin{adjustbox}{width=0.99\textwidth}
\begin{tabular}{r|c|c|c}
& Reference & SIII & SIV\\
\toprule
{\rotatebox{90}{CV}}&\includegraphics[width=0.3\textwidth]{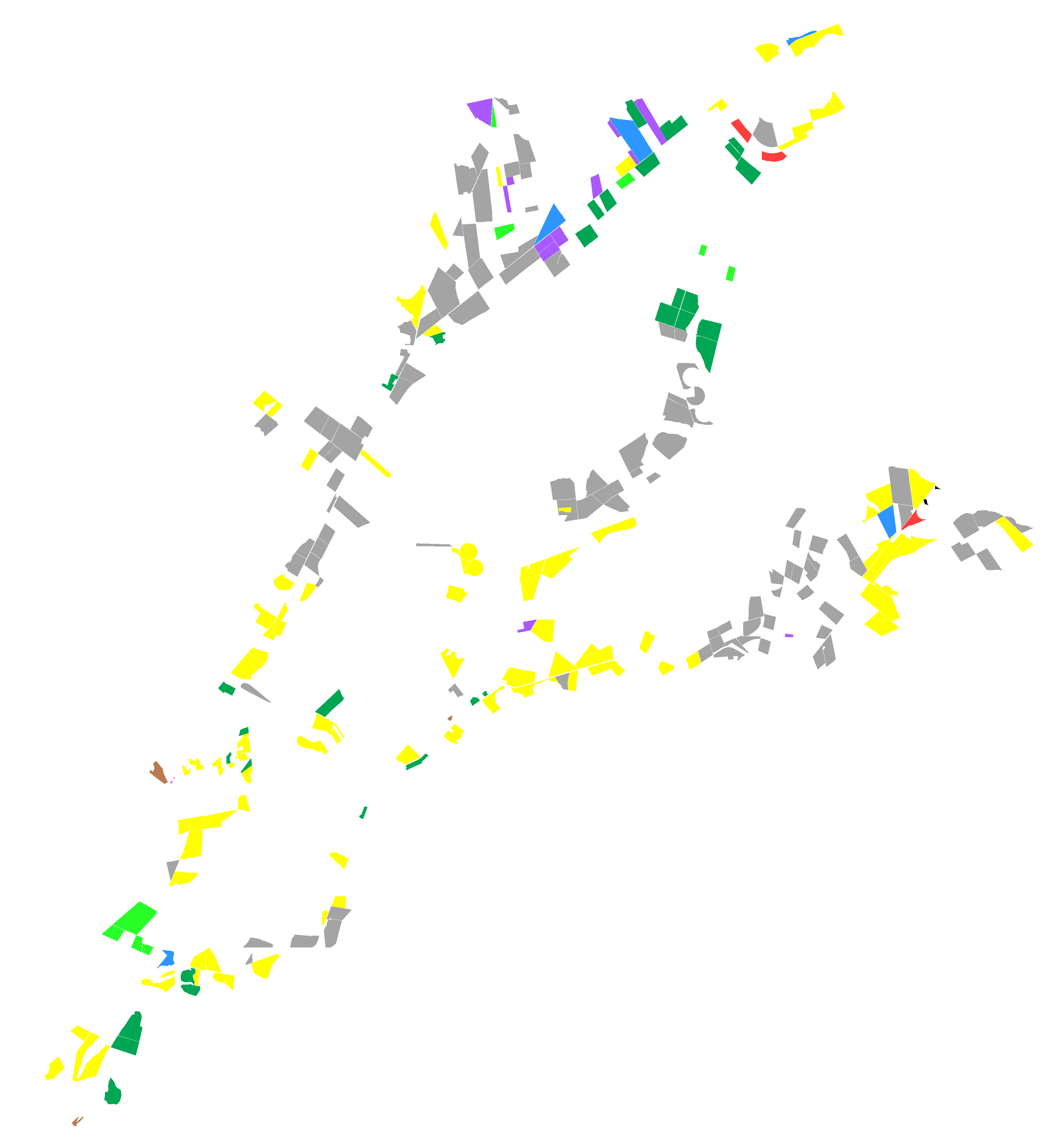}& \includegraphics[width=0.3\textwidth]{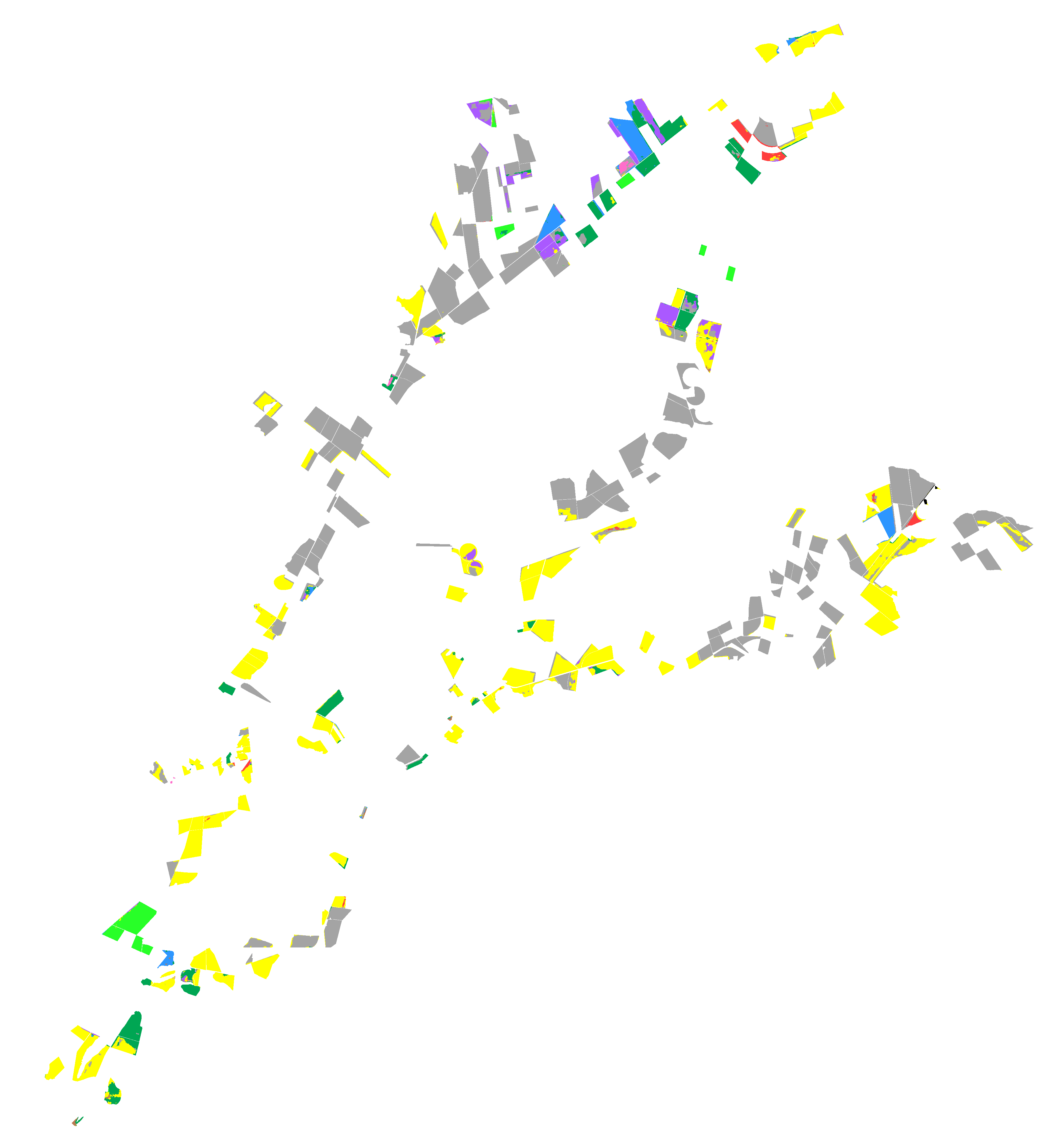}& \includegraphics[width=0.3\textwidth]{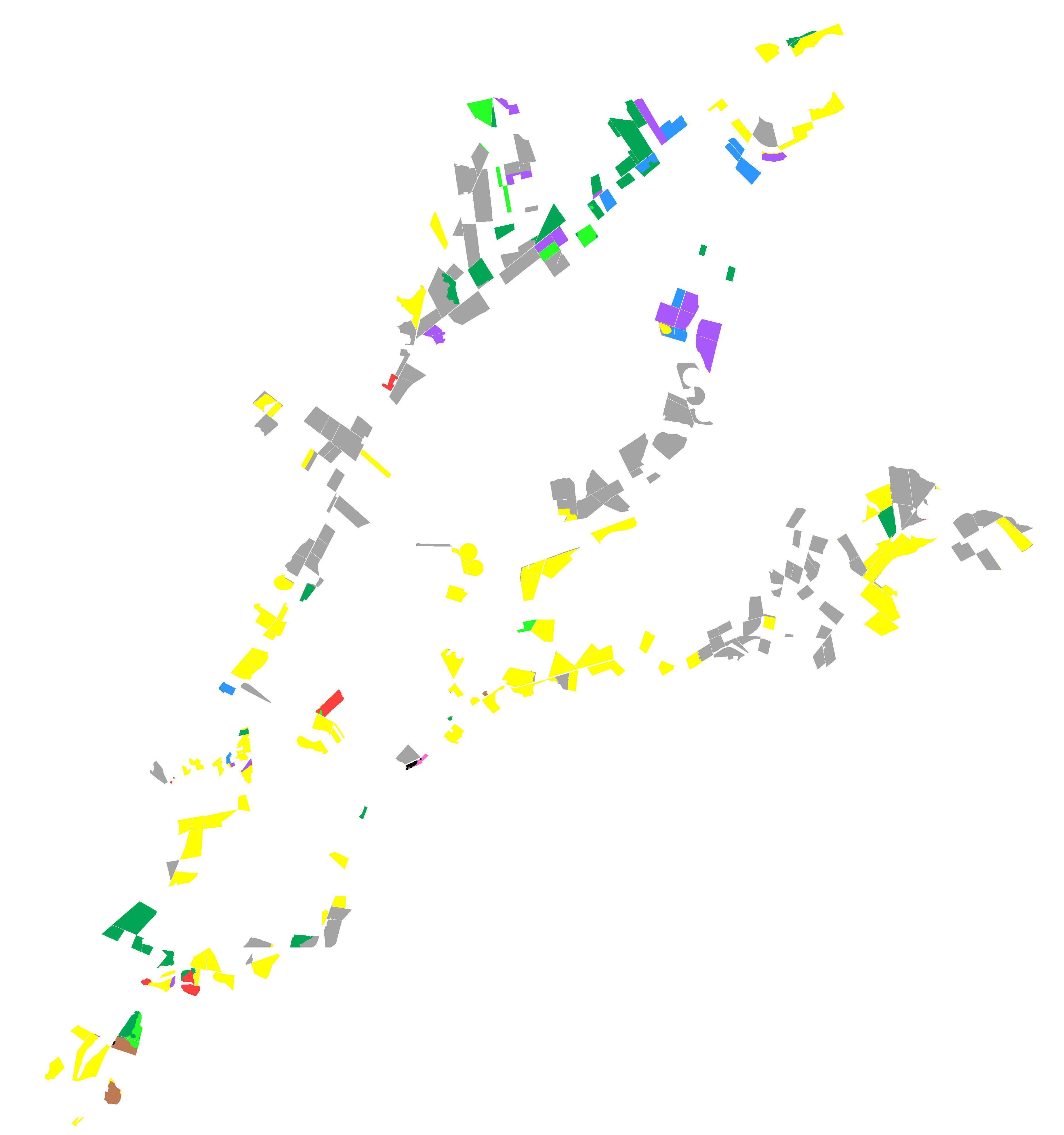}\\\midrule
{\rotatebox{90}{LEM}}&\includegraphics[width=0.3\textwidth]{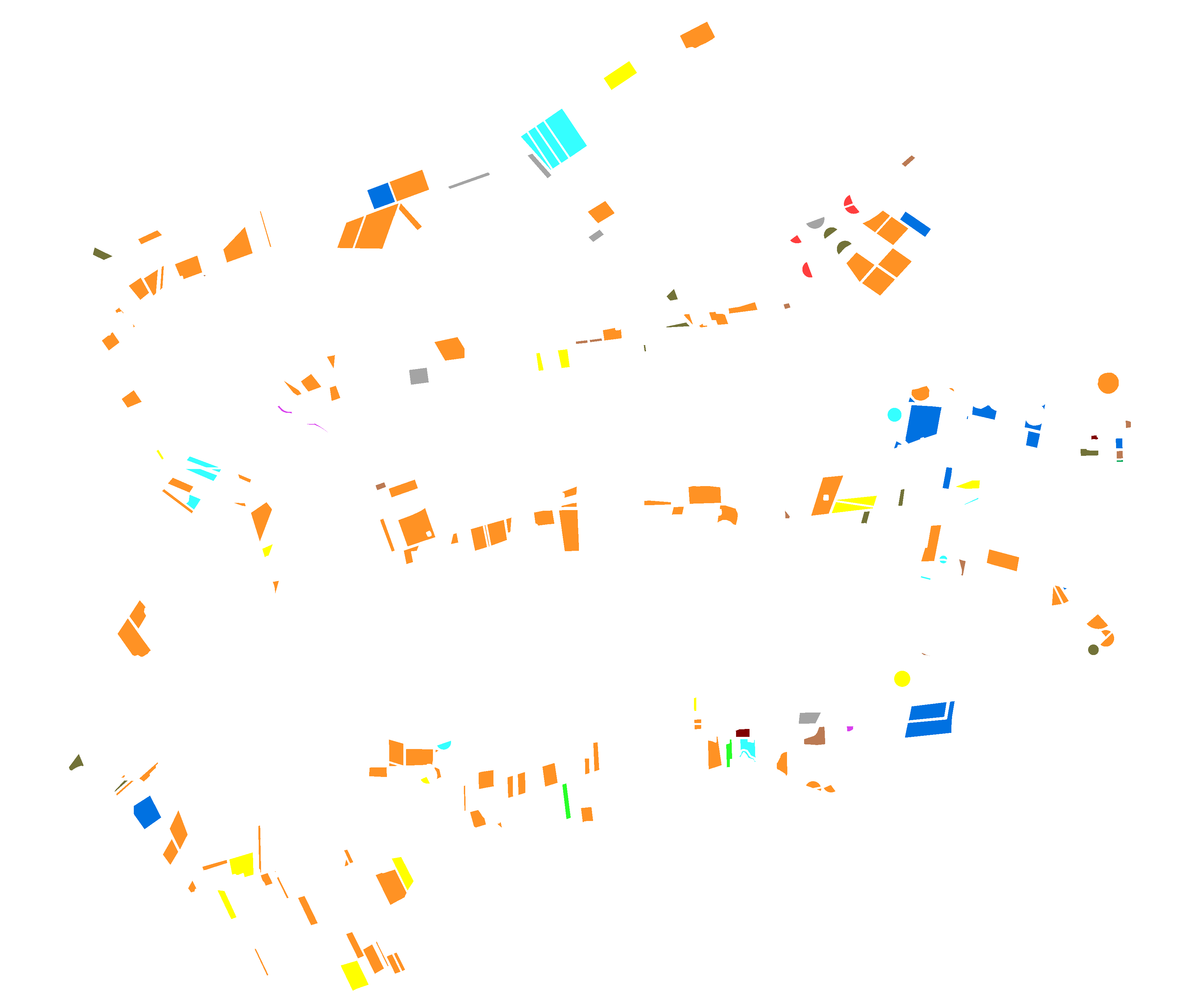}& \includegraphics[width=0.3\textwidth]{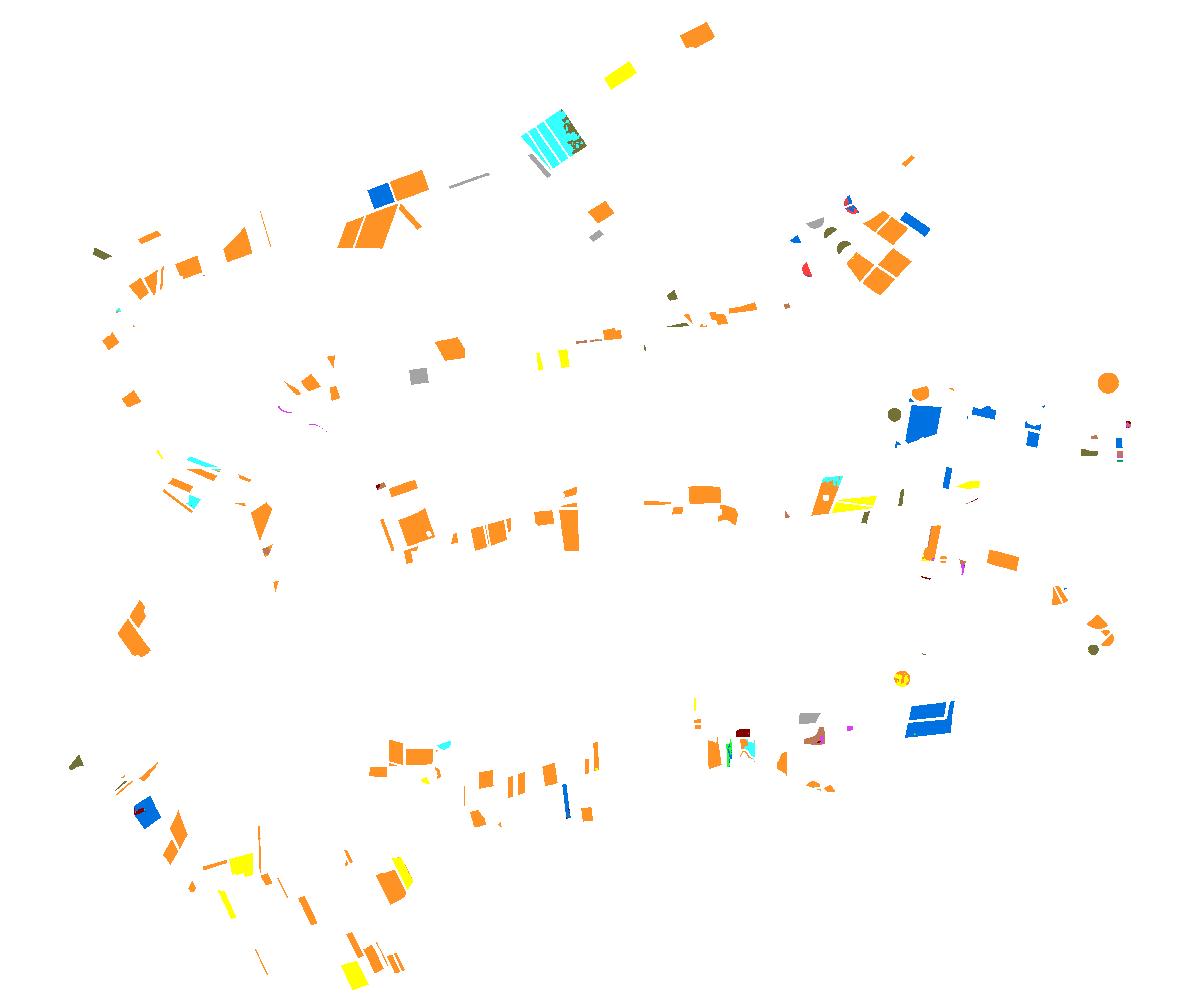}& \includegraphics[width=0.3\textwidth]{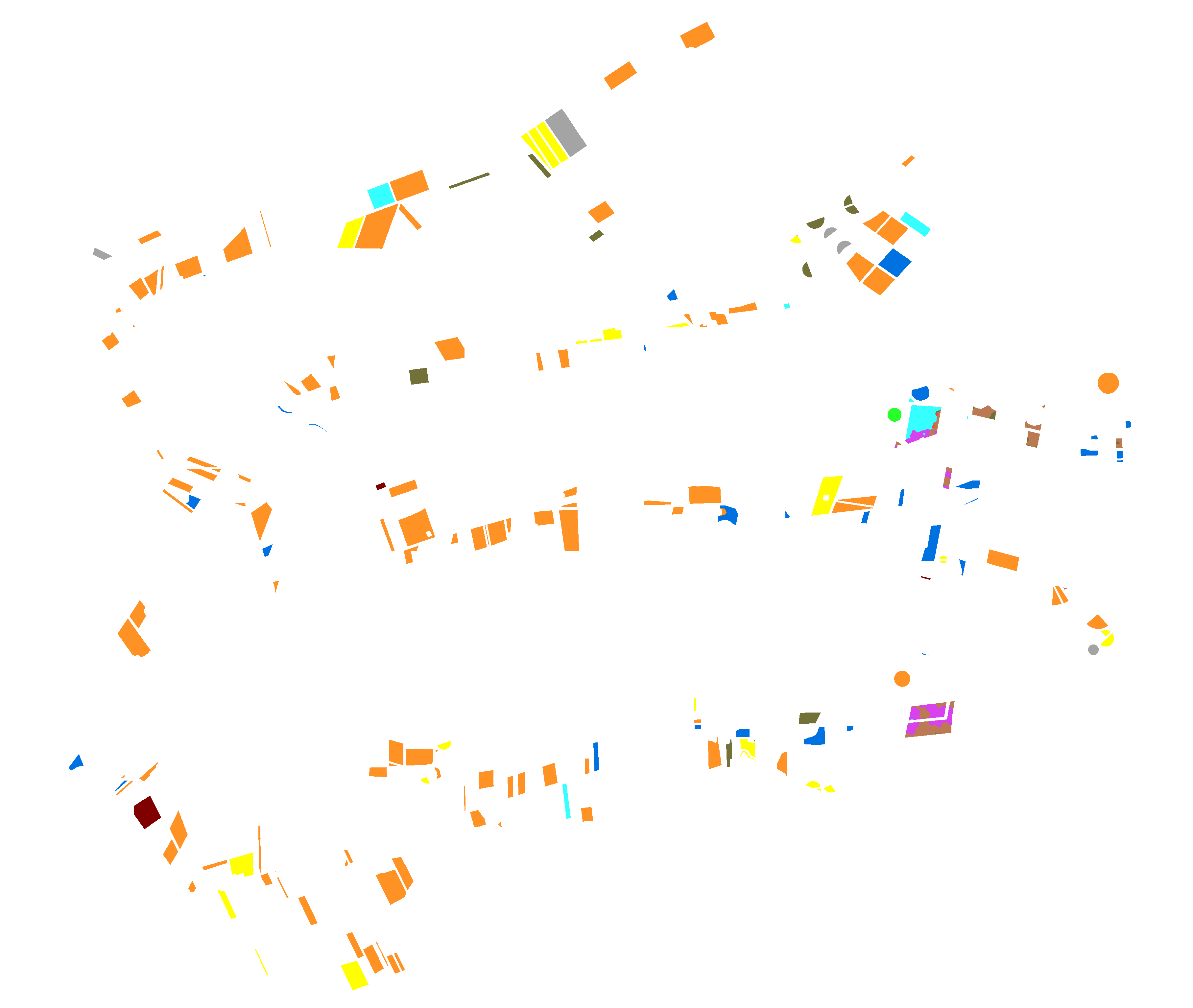}\\\bottomrule
\end{tabular}
\end{adjustbox}
\caption{Maps of the class output for scenarios SIII and SIV for the test region for the CV and LEM datasets. Same color legend as in Fig.~\ref{tab:clips1s2}.}
\label{tab:clips3s4}
\end{figure*}

Finally, the qualitative results for scenarios SIII and SIV are presented in Fig.~\ref{tab:clips3s4}. As expected, the SIII classification map is remarkably better than the SIV map for both the CV and LEM datasets, especially for the minor classes. 

Overall, our approach was able to identify correctly the major crop types whose global proportions were different and representative in the dataset. Using the exact proportion in each dataset for the major crops, it achieved an accuracy of more than 90\% for the target regions. Using global class proportions estimated from census data, the trained classifier could identify the actual crop types at each pixel location with relatively high accuracies (80\% and 85\% for the target regions). Given the exact proportions for each bag and with small bag sizes, it achieved results close to those of a fully supervised approach, even for minor classes. The approach failed to identify minor classes with very small or similar global proportions, which indicates that to train a model with a unique vector of global proportions, the sampling for each bag needs to follow the same proportions. This is impossible to guarantee in highly unbalanced datasets, even for large bag sizes. Nonetheless, this problem is not restricted to our approach, and commonly used solutions for tackling data imbalance, such as a weighted categorical cross-entropy or focal loss function, could easily be incorporated into our method. Finally, as we hypothesized, increasing the bag size gave the best performance, especially for more representative classes, when considering the global proportions.

\section{Conclusions}\label{sec:conclusion}

This work proposes an end-to-end weakly supervised contrastive-learning method based on proportions to discriminate among crop types from large-scale RS imagery. The methodology uses the input raw data and class proportions as priors to implement a DL algorithm that jointly performs contrastive clustering and feature learning in an end-to-end fashion constrained to known class proportions for the target agricultural area. Despite being trained without ground truth labels at the pixel level, our approach can deliver pixel-wise classification, closing the gap between supervised and weakly supervised models for crop classification using RS images.  

We tested our methodology with two large-scale agriculture datasets from Brazil. Information about the proportions of crops was available from government census data. We did extensive experiments using optical and SAR images to assess the ability of the model to deal with different data sources in single-date and multi-temporal settings, respectively. 

Our results show that in a real-life scenario with access to the exact crop-type proportions for a given dataset, the method reached accuracies above 90\% for the major crop types. On the other hand, with access only to the estimated crop proportions for a given agricultural region, it can discriminate among the different crop types with an accuracy of above 80\%. However, our methodology also struggles to identify minor classes based solely on global proportions, which seems to be related to the well-known class imbalance problem. We also assessed the sensitivity of the model to the bag size and concluded that the model would perform better for small bag sizes when working with the exact bag class proportions. In contrast, considering global proportions, the performance of the model improved as the bag size increased. 

The success of LLP-Co demonstrates that it offers a powerful alternative for using prior information to design weakly supervised methods for agricultural applications and other research areas, such as forestry and climate, for which census information is available at different levels. While this work explored the potential of using proportions efficiently for crop classification in agriculture, future work may include known efficient methods for dealing with highly unbalanced datasets to improve the identification of minor classes. In addition, we plan to explore the sensitivity of the model to some of the hyperparameters, such as the batch size, and we also intend to investigate various augmentation strategies. 

\bibliographystyle{IEEEtran}
\bibliography{llp_crop_main.bib}
 
\end{document}